\documentclass[letterpaper]{article} 
\usepackage{aaai21}  

\usepackage[utf8]{inputenc}
\usepackage{xcolor}
\usepackage{amsmath}
\usepackage{amsfonts}
\usepackage{amsthm}
\usepackage{amsxtra}
\usepackage{amssymb}
\usepackage{soul}
\usepackage[nameinlink]{cleveref}
\usepackage[numbers]{natbib}
\usepackage{algorithm}
\usepackage{algorithmic}
\usepackage{graphicx}
\usepackage{xspace}
\usepackage{lipsum}
\usepackage{enumerate}
\usepackage{subcaption}
\usepackage{booktabs}
\usepackage{multicol}
\usepackage[switch]{lineno}

\newcommand{\E}{\mathbb{E}}
\newcommand{\D}{\mathcal{D}}

\newcommand{\KL}[2]{\operatorname{KL}(\,{\textstyle#1}\,\|\,{\textstyle#2}\,)}
\DeclareMathOperator{\tr}{tr}

\newcommand{\xs}{x^\star}
\newcommand{\e}{\varepsilon}
\newcommand{\eff}{\text{eff}}
\newcommand{\tbf}[1]{\textbf{#1}}

\DeclareMathOperator*{\argmin}{argmin}
\DeclareMathOperator*{\argmax}{argmax}

\newcommand{\norm}[1]{\left\|#1\right\|}

\newtheorem{thm}{Theorem}[section]
\newtheorem{proposition}[thm]{Proposition}
\newtheorem{lemma}[thm]{Lemma}

\newtheorem{definition}[thm]{Definition}
\theoremstyle{definition}
\newtheorem{example}[thm]{Example}

\newcommand{\rob}{\textbf{Rob}}
\newcommand{\sta}{\textbf{St}}
\newcommand{\ct}{C-10}
\newcommand{\co}{C-100}
\newcommand{\fmn}{F-MNIST}
\newcommand{\mn}{MNIST}
\newcommand{\sv}{SVHN}
\newcommand{\im}{IMG}
\newcommand{\air}{Aircraft}
\newcommand{\bir}{Birds}
\newcommand{\ca}{Cars}
\newcommand{\fl}{Flowers}
\newcommand{\dog}{Dogs}
\newcommand{\ind}{Indoor}

\newcommand{\mt}[1]{{\color{black}#1}}
\newcommand{\bff}[1]{{\textbf{#1}}}

\title{Adversarial Training Reduces Information and Improves Transferability}

  \author{%
 Matteo Terzi\textsuperscript{\rm 1},
 Alessandro Achille\textsuperscript{\rm 2},
 Marco Maggipinto\textsuperscript{\rm 1},
 Gian Antonio Susto\textsuperscript{\rm 1}
 \\
}

\affiliations{
\textsuperscript{\rm 1}Department of Information Engineering\\
University of Padova\\
\{terzimat,marco.maggipinto\}@dei.unipd.it, gianantonio.susto@unipd.it \\
\textsuperscript{\rm 2}Department of Computer Science\\
  University of California, Los Angeles\\
  achille@cs.ucla.edu
}

\graphicspath{{./fig/}}

\begin{document}
\maketitle

\begin{abstract}
Recent results show that features of adversarially trained networks for classification, in addition to being robust, enable desirable properties such as invertibility. 
The latter property may seem counter-intuitive as it is widely accepted by the community that classification models should only capture the minimal information (features) required for the task. 
Motivated by this discrepancy, we investigate the dual relationship between Adversarial Training and Information Theory. We show that the Adversarial Training can improve linear transferability to new tasks, from which arises a new trade-off between transferability of representations and accuracy on the source task. We validate our results employing robust networks trained on CIFAR-10, CIFAR-100 and ImageNet on several datasets. 
Moreover, we show that Adversarial Training reduces Fisher information of representations about the input and of the weights about the task, and we provide a theoretical argument which explains the invertibility of deterministic networks without violating the principle of minimality. Finally, we leverage our theoretical insights to remarkably improve the quality of reconstructed images through inversion.
\end{abstract}

\section{Introduction}
In the last 10 years, Deep Neural Networks (DNNs) dramatically improved the performance in any computer vision task. However, the impressive accuracy comes at the cost of poor robustness to small perturbations, called {\em adversarial perturbations}, that lead the models to predict, with high confidence, a wrong class~\cite{goodfellow2014explaining,terzi2020directional}.
This undesirable behaviour led to a flourishing of research works ensuring robustness against them. State-of-the-art approaches for robustness are provided by Adversarial Training (AT)~\cite{madry2017towards} and its variants~\cite{zhang2019theoretically}. 
The rationale of these approaches is to find worst-case examples and feed them to the model during training or constraining the output to not change significantly under small perturbations.
However, robustness is achieved at the expense of a decrease in accuracy: the more a model is robust, the lower its accuracy will be~\cite{tsipras2018robustness}. 
This is a classic “waterbed effect” between precision and robustness ubiquitous in optimal control and many other fields.
Interestingly, robustness is not the only desiderata of adversarially trained models: their representations are semantically meaningful and they can be used for other Computer Vision (CV) tasks, such as generation and (semantic) interpolation of images.
More importantly, AT enables invertibility, that is the ability to reconstruct input images from their representations~\cite{ilyas2019adversarial} by solving a simple optimization problem. 
This is true also for {\em out-of-distribution} images meaning that robust networks do not {\em destroy} information about the input. Hence, how can we explain that, while robust networks preserve information, they lack in generalization power?

In this context, obtaining good representations for a task has been the subject of {\em representation learning} where the most widely accepted theory is Information Bottleneck (IB)~\cite{tishby2000information,alemi2016deep,achille2018information} which calls for reducing  information in the activations, arguing it is necessary for generalization.
More formally, let $x$ be an input random variable and $y$ be a target random variable, a good representation $z$ of the input should be maximally expressive about for $y$ while being as concise as possible about $x$. 
The solution of the optimal trade-off can be found by optimizing the Information Lagrangian:
$$
\min_{z} -I (z,y) + \beta I(z,x)
$$
where $\beta$ controls how much information about $x$ is conveyed by $z$. Both AT and IB at their core aim at finding good representations: the first calls for representations that are {\em robust} to input perturbations while the latter finds {\em minimal} representations sufficient for the task. 
How are these two methods related? Do they share some properties?
More precisely, does the invertibility property create a contradiction on IB theory? In fact, if generalization requires discarding information in the data that is not necessary for the task, it should not be possible to reconstruct the input images.

Throughout this paper we will (i) investigate the research questions stated above, with particular focus on the connection between IB and AT and as a consequence of our analysis, (ii) we will reveal new interesting properties of robust models.

\subsection{Contributions and related works}
A fundamental result of IB is that, in order to generalize well on a task, $z$ has to be sufficient and minimal, that is, it should contain only the information necessary to predict $y$, which in our case is a target class. Apparently, this is in contradiction with the evidence that robust DNNs are invertible maintaining almost all the information about the input $x$ even if is {\em not} necessary for the task.
However, what matters for generalization is not the information in the activations, but information in the weights (PAC-Bayes bounds)~\cite{achille2019information}. Reducing information in the weights, yields to reduction in the {\em effective} information in the activations at test time. Differently from IB theory, ~\cite{achille2019information} claims that the network does not need to destroy information in the data that is not needed for the task: it simply needs to make it inaccessible to the classifier, but otherwise can leave it lingering in the weights.
That is the case for ordinary learning. 
As for AT, robustness is obtained at cost of lower accuracy on natural images~\cite{madry2017towards,tsipras2018robustness}, suggesting that only the robust features are extracted by the model~\cite{ilyas2019adversarial}: How can be this conciliated with invertibility of robust models?
This paper shows that, while AT {\em preserves} information about the data that is irrelevant for the task in the weights (to the point where the resulting model is invertible), the information that is {\em effectively} used by the classifier does not contain all the details about the input $x$. In other words, the network is not effectively invertible: what really matters is the accessible information stored in the weights. 
In order to visualize this fact, we will introduce {\em effective images}, that are images that represent what the classifier "sees". Inverting learned representations is not new, and it was solved in ~\cite{mahendran2015understanding,yosinski2015understanding, ulyanov2018deep, kingma2013auto}; however, these methods either inject external information through priors or explicitly impose the task of reconstruction contrary to robust models.

The main contribution of this work can be summarized as follows.
If representations contain all the information about the input $x$, then adversarially trained models should be better at transfering features to different tasks, where aspects of the data that were irrelevant to the task it was (pre)-trained on were neither destroyed nor ignored, but preserved. 
To test this hypothesis, we perform linear classification (fine-tune the last layer) for different tasks. We show that AT improves linear transferability of deep learning models across diverse tasks which are {\em sufficiently different} from the source task/dataset.
Specifically, the farther two tasks are (as measured by a task distance), the higher the performance improvement that can be achieved by training a linear classifier using an adversarially-trained model (feature, or backbone) compared to an ordinarily trained model. 
Related to this, in \cite{shafahi2019adversarially} the transferability of robustness to new tasks is studied experimentally; differently, in the present work we study the linear transferability of natural accuracy. Moreover, we also analytically show that, confirming empirical evidence~\cite{ilyas2019adversarial}, once we extract robust features from a backbone model, {\em all} the models using these features have to be robust. 

We will also show that adversarial regularization is a lower-bound of the regularizer in the Information Lagrangian, so AT in general results in a loss of accuracy for the task at hand. 
The benefit is increased transferability, thus showing a classical tradeoff of robustness (and its consequent transferability) and accuracy on the task for which it is trained. This is a classic "waterbed effect" between precision and robustness ubuiquitous in optimal control. Regarding the connection with IB, we show analytically that AT reduces the effective information in the activations about the input, as defined by~\cite{achille2019information}. Moreover, we show empirically that adversarial training also reduces information in the weights and its consequences.

Finally, we show that injecting effective noise once during the inversion process dramatically improves reconstruction of images in term of convergence and quality of fit. 

\mt{In order to facilitate the reading, the manuscript is organized as follows. ~\Cref{sec:notation} provides the necessary notation. \Cref{sec:at_reduce} presents all the theoretical building blocks by showing the connection between AT and IB. Based on the previous results,~\Cref{sec:invertibility} shows why there is no contradiction between minimality of representations and invertibility of robust models, and \Cref{sec:transfer} shows that robust features can transfer better to new tasks.}

\section{Preliminaries and Notation}
\label{sec:notation}
We introduce here the notation used in this paper. We denote a dataset of $N$ samples with $\D = \{(x_i, y_i)\}_{i=1}^N$ where $x\in X$ is an input, and $y\in Y$ is the target class in the finite set $Y = \{1, \dots, K\}$. More in general, we refer to $y$ as a random variable defining the "task".
In this paper we focus on classification problems using cross-entropy  $L_\D(w) = \E_{(x,y) \sim \D}[\ell (x,y; w)]$ on the training set $\D$ as objective where $\ell (x,y; w) = -\log p_w(y|x)$ and $p_w(y|x)$ is encoded by a DNN. 
The loss $L_\D(w)$ is usually minimized using \textit{stochastic gradient descent} (SGD)~\cite{bottou2018optimization}, which updates the weights $w$ with a noisy estimate of the gradient computed from a mini-batch of samples. 
Thus, weights update can be expressed by a stochastic diffusion process with non-isotropic noise~\cite{li2017stochastic}.
In order to measure the (asymmetric) dissimilarity between distributions we use the Kullbach-Liebler divergence between $p(x)$ and $q(x)$ given by $\KL{p(x)}{q(x)} = \E_{x\sim p(x)}\big[\log (p(x)/q(x))\big]$. 
It is well-known that the second order approximation of the KL-divergence is $\E_x\KL{p_w(y|x)}{p_{w+\delta w}(y|x)} = \delta w^t F \delta w + o(\|\delta w\|^2)$
where $F$ is the \textit{Fisher Information Matrix} (FIM), defined by $F = \E_{x, y\sim p(x)p_w(y|x)}[\nabla \log p_w(y|x)\nabla \log p_w(y|x)^t] = \E_{x\sim p(x)p_w(y|x)}[- \nabla^2_w \log p_w(y|x)].$ 
The FIM gives a local measure of how much a perturbation $\delta w$ on parameters $w$, will change $p_w (y|x)$ with respect to KL divergence~\cite{martens2014new}.
Finally, let $x$ and $z$ be two random variables. The \textit{Shannon mutual information} is defined as $I(x; z) = \E_{x \sim p(x)}[\KL{p(z|x)}{p(z)}]$. 
Throughout this paper, we indicate the {\em representations} before the linear layer as $z = f_w(x)$, where $f_w(x)$ is called {\em feature extractor}.

\paragraph{Adversarial Training}
AT aims at solving the following min-max problem:
\begin{equation} \label{eq:at}
\begin{cases}
\min_w \E_{(x,y) \sim \D}[\ell (\xs,y; w)] \\
\delta^\star = \argmax_{\|\delta\|_2 < \e} \ell (x + \delta, y; w) \\
\xs = x + \delta^\star
\end{cases}
\end{equation}
In the following we denote $\E_{(x,y) \sim \D}[\ell (\xs,y; w)]$ with $L_\D^\star(w)$. 
We remark that by $\E_{(x,y) \sim \D}$ we mean the empirical expectation over $N$ elements of the dataset. 
Intuitively, the objective of AT is to ensure stability to small perturbations on the input. With cross-entropy loss this amounts to require that $\KL{p_w(x+\delta)}{p_w(x)} \le \gamma$, with $\gamma$ small.
Depending on $\varepsilon$, we can write~\Cref{eq:at} as:
\begin{equation}
\begin{split}
\min_w \E_{(x,y) \sim \D}[\ell (x,y; w)] + \\
\beta \max_{\norm{\delta}_2 \le \varepsilon}{\KL{p_w(y|x+\delta)}{p_w(y|x)}}
\end{split}
\end{equation}
which is the formulation introduced in~\cite{zhang2019theoretically} when using cross-entropy loss. We define the (weak) inversion of features as:
\begin{definition}[Inversion] Let $\bar{z} = f_w(x)$ be the final representation (before linear classifier) of an image $x$, and let $f_w$ be the robust feature extractor. 
The reconstructed image (inversion) is the solution of the following problem: 
\begin{equation}
\label{eq:inversion}
\hat{x} (x; z) = f^{-1}_w (\bar{z}) = \argmin_{x'}{\norm{\bar{z} - f_w(x')}_2} 
\end{equation}
where the initial condition of $x'$ is white noise $x' (0)\sim N(0.5,\sigma)$, where $\sigma$ is the noise scale.
\end{definition}

\begin{figure*}[htbp]
    \centering
    \includegraphics[width=\textwidth]{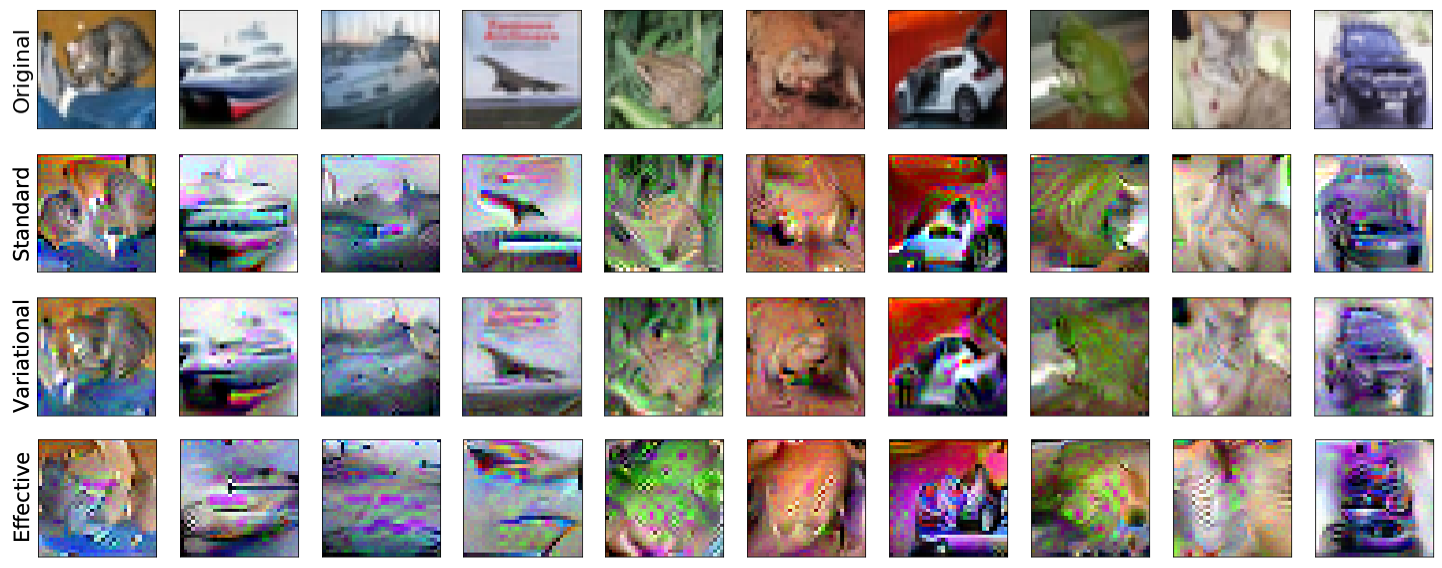}
    \caption{Inversion using the standard and variational ResNet-50 model. (Top row) Original images. (Second row) Images reconstructed optimizing~\Cref{eq:inversion}. (Third row) Images reconstructed by adding noise only once. (Bottom row) Effective images obtained optimizing ~\Cref{eq:effective_image}.}
    \label{fig:inv_all_model}
\end{figure*}

\section{AT reduces information} 
\label{sec:at_reduce}
In this section, we analytically show why a robust network, even if it is invertible at test time, is effectively not invertible as a consequence of noise injected by SGD. 
We first define the Fisher $F_{z\mid x}$ of representations w.r.t. inputs.
\begin{definition}
The FIM $F_{z\mid x}$ of representations w.r.t the input distribution is defined as:
\begin{align}
\label{eq:fim_rep}
    F_{z|x} &= \E_{x\sim p (x)} \E_{z\sim p_w(z|x)} {\nabla_x \nonumber \log p_w (z|x)} {\nabla_x \log p_w (z|x)}^t \\
    &= \E_{x\sim p (x)} S(z|x)
\end{align}
where $S(z|x)$ is the sensitivity matrix of the model at a fixed input location $x$.
\end{definition}

In the next proposition we relate AT to \Cref{eq:fim_rep}, showing that, requiring the stability of $f_w$ w.r.t. $x$ is equivalent to regularize the FIM $F_{z|x}$.

\begin{proposition}
\label{prop:kl_adv}
Let $\delta \in X$ be a small perturbation such that $\norm{\delta}_2 = \varepsilon$.\footnote{We would like to note that the practical implementation only requires $\norm{\delta}_2 \le \varepsilon$. However, in practice, it is possible to see that for small $\varepsilon$, the norm of $\delta$ is almost always $\varepsilon$.} Then,
\begin{equation}\label{eq:kl_approx}
\max_{\norm{\delta}_2} \KL{p_w(z|x+\delta)}{p_w(z|x)} \approx \dfrac{\varepsilon^2}{2} v_{\lambda_1}^t{S(z|x)}v_{\lambda_1}
\end{equation}
where $v_{\lambda_1}$ is the (unit-norm) eigen-vector corresponding to the first principal eigenvalue $\lambda_1$.
\end{proposition}
Hence, AT is equivalent to regularize the Fisher of representation $z$ with respect to inputs $x$.
By applying white Gaussian noise instead of adversarial noise,~\Cref{eq:kl_approx} would become $\KL{p_w(z|x+\delta)}{p_w(z|x)} \approx \dfrac{\varepsilon^2}{2n} \tr{S(z|x)}$, where $n$ is the input dimension. It is easy to see that $\dfrac{\tr{S(z|x)}}{n} \leq  v_{\lambda_1}^t{S(z|x)}v_{\lambda_1}$, meaning that Gaussian Noise Regularization (GNR) is upper bounded by AT: the inefficiency of GNR increases as the input dimension increases, causing that many directions preserve high curvature. 
~\cite{tsipras2018robustness} showed that AT, for a linear classification problem with hinge loss, is equivalent to penalize the $\ell_2$-norm of weights. The next example shows that when using cross-entropy loss, penalizing the Fisher $F_{z|x}$ yields a similar result.
\begin{example}[Binary classification] Assume a binary classification problem where $y \in \{-1,1\}$ Let
$p(y=1|x) = 1 - p(y=-1|x) = \texttt{sigmoid} (w^tx)$. 
Then we have:
\begin{equation*}
\begin{split}
    F_{z|x} = c ww^t \;, \;\tr(F_{z|x}) =
    c\norm{w}_2^2, \; c = \E_x[p(-1 | x) p(1 |x)]
\end{split}
\end{equation*}
\end{example}

\begin{figure*}
\centering
\begin{minipage}{.32\textwidth}
    \vspace{0pt}
    \centering
    \includegraphics[width=1\textwidth, height=3.12cm]{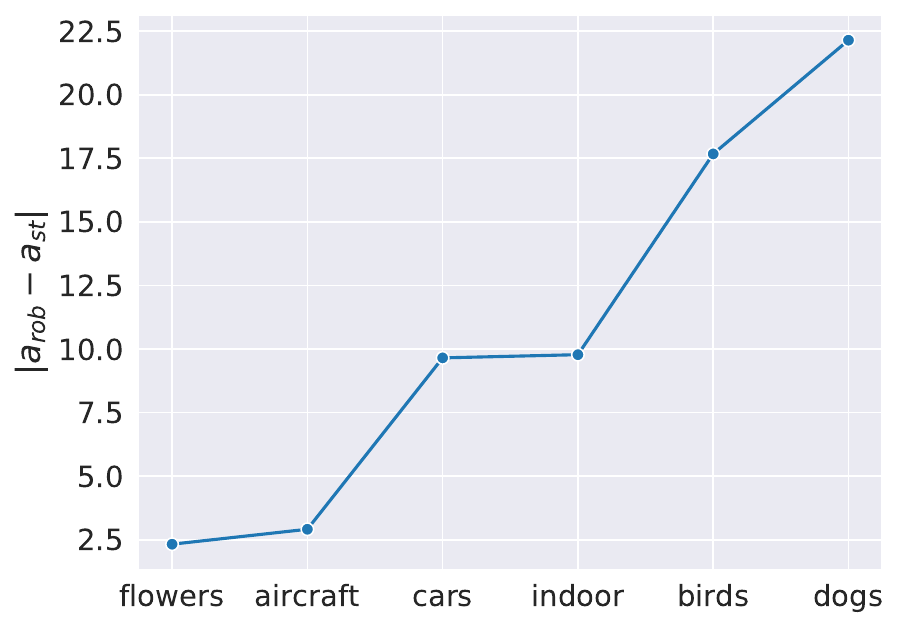}
  \captionof{figure}{Accuracy gap between the robust and standard model as the distance from the source task increases.}
  \label{fig:img_acc_order}
\end{minipage}%
\hspace{2pt}
\begin{minipage}{.32\textwidth}
  \centering
\includegraphics[width=1\textwidth, height=3.3cm]{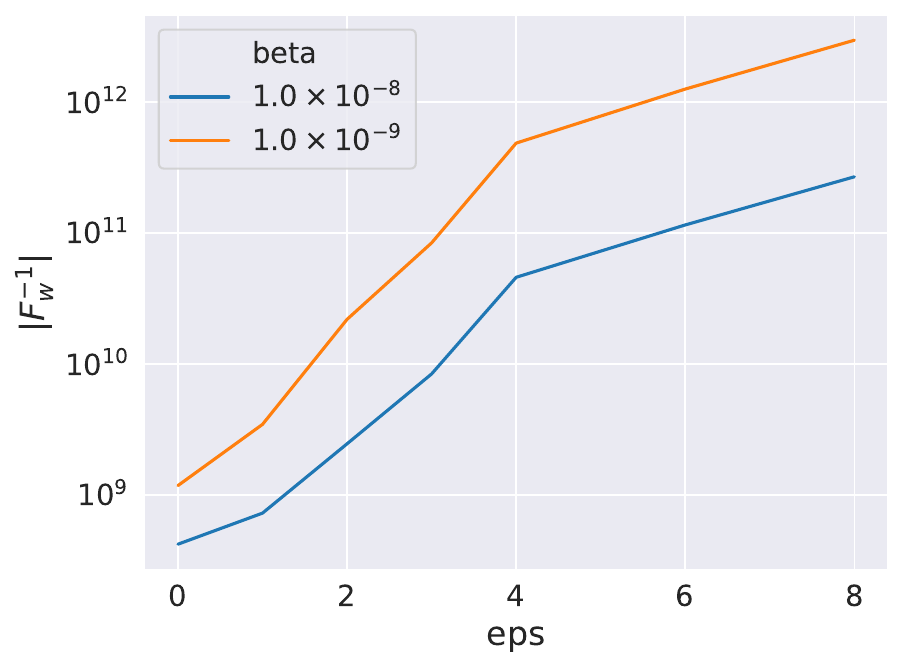}
    \caption{Flatness of Fisher information as measured by the norm of embedding (log-variance).}
  \label{fig:fisher_cifar}
\end{minipage}
\hspace{2pt}
\begin{minipage}{.32\textwidth}
  \centering
\includegraphics[width=1\textwidth, height=3.2cm]{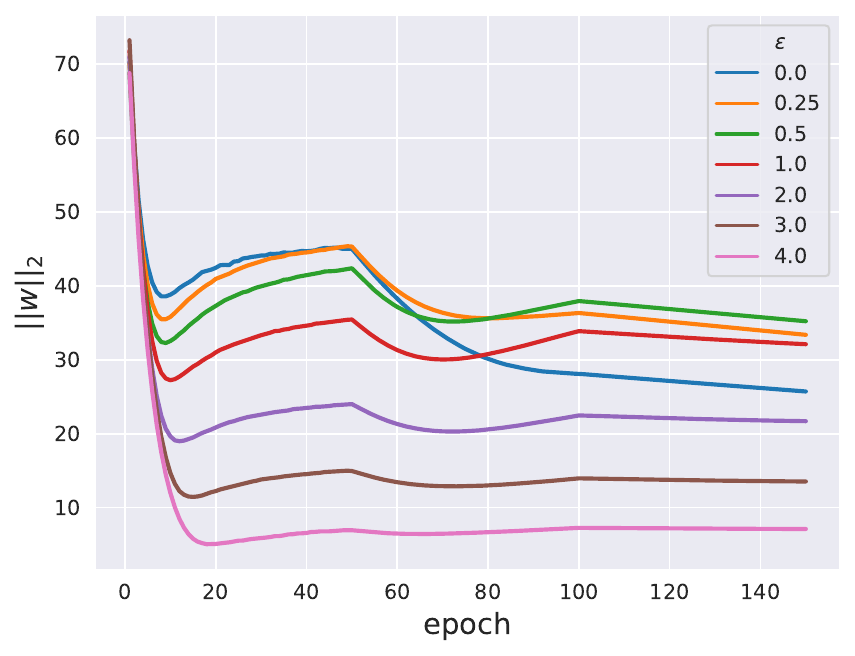}
  \captionof{figure}{Norm of weights for different $\varepsilon$. Robust and standard training differ in the dynamics of $\norm{w}_2$.}
  
  \label{fig:weights_norm}
\end{minipage}
\end{figure*}

The previous example may suggest that with $\ell_2$-perturbations AT may reduce the $l_2$-norm of the weights. We trained robust models with different $\varepsilon$ (with the same seed) to verify this claim: as reported in~\Cref{fig:weights_norm}, we discovered that it is true only for $\varepsilon > 1$, pointing out that there may exist two different regimes.





What we are interested in is the relation between the Shannon Mutual Information $I(z,x)$ and the Fisher Information in the activations $F_{z|x}$. However, in adversarial training there is nothing that is stochastic but SGD. For this reason, ~\cite{achille2019information} introduced \emph{effective information}. 
The idea under this definition is that, even though the network is deterministic at the end of training, what matters is the noise that SGD injects to the classifier. 
Thus, the effective information is a measure of the information that the network effectively uses in order to classify. Before continuing, we need to quantify this noise applied to weights.

\begin{definition}[Information in the Weights]
\label{def:complexity-dnn}
    The complexity of the task $\D$ at level $\beta$, using the posterior $Q(w|\D)$ and the prior $P(w)$, is
    \begin{equation}
    \label{eq:kl-complexity}
    \begin{split}
            C_\beta(\D; P, Q) = \E_{w \sim Q(w|\D)}[L_\D(p_w(y|x))]
            + \\ \beta  \underbrace{\KL{Q(w|\D)}{P(w)}}_\text{Information in the Weights},
    \end{split}
    \end{equation}
    where $\E_{w \sim Q(w|\D)}[L_\D(p_w(y|x))]$ is the (expected) reconstruction error of the label under the ``noisy'' weight distribution $Q(w|\D)$; $\KL{Q(w|\D)}{P(w)}$ measures the entropy of $Q(w|\D)$ relative to the prior $P(w)$. If $Q^*(w|\D)$ minimizes \Cref{eq:kl-complexity} for a given $\beta$, we call $\KL{Q^*(w|\D)}{P(w)}$ the \emph{Information in the Weights} for the task $\D$ at level $\beta$.
\end{definition}
Given the prior $P(w) \sim N(0, \lambda^2 I)$, the solution of the optimal trade-off is given by the distribution $Q(w|\D) \sim N(w^\star, \Sigma^\star )$ such that $\Sigma^\star = \dfrac{\beta}{2}\left(F_w + \dfrac{\beta}{2\lambda^2} I \right)^{-1}$ with $F_w  \approx \nabla^2_w L_\D(w)$. 
 The previous definition tells us that if we perturb uninformative weights, the loss is only slightly perturbed. This means that information in the activations that is not preserved by such perturbations is not used by the classifier.

\begin{definition}(Effective Information in the Activations \cite{achille2019information}).\label{def:eff_info}
Let $w$ be the weights, and let $n \sim N(0, \Sigma^*_w)$, with $\Sigma^*_w = \beta F^{-1}(w)$ be the optimal Gaussian noise minimizing \Cref{eq:kl-complexity} at level $\beta$ for a prior $N(0, \lambda^2 I)$. We call \emph{effective information} (at noise level $\beta$) the amount of information about $x$ that is not destroyed by the added noise:
\begin{equation}
\label{eq:effective-information}
I_{\eff,\beta}(x;z) = I(x;z_n),
\end{equation}
where $z_n = f_{w + n}(x)$ are the activations computed by the perturbed weights $w+n \sim N(w, \Sigma^*_w)$.
\end{definition}

By Prop. 4.2(i) in~\cite{achille2019information} we have that the relation between $F_{z|x}$ and effective information is given by:
\begin{equation}
    \label{eq:shannon-fisher}
        I_{\text{eff},\beta}(x;z) \approx H(x) - \E_x\Big[\frac{1}{2} \log\Big(\frac{(2\pi e)^k}{|F_{z|x}|}\Big)\Big],
\end{equation}
where $H(x)$ is the entropy of input distribution. ~\Cref{eq:shannon-fisher} shows that AT compresses data similarly to IB. With AT, the noise is injected in the input $x$ and {\em not} only in the weights.
In order to reduce the \emph{effective} information that the representations have about the input (relative to the task), it is sufficient to decrease $|F_{z|x}|$, that is, increasing $\varepsilon$. In the Supplementary Material, we show how details about $x$ are discarded varying $\varepsilon$.


\vspace{-3mm}
\paragraph{AT reduces the information in the weights}
We showed that AT reduces effective information about $x$ in the \emph{activation}. 
However,~\cite{achille2019information} showed that to have guarantees about generalization and invariance to nuisances at test time one has to control the trade off between sufficiency for the task and information the weights have about the dataset. A natural question to ask is whether reducing information in the activations implies reducing information in the weights, that is the mutual information $\beta I(w; D)$ between the weights and the dataset. The connection between weights and activation is given by the following formula (~\Cref{prop:re-emergence}):
\begin{equation}
F_{z|x} = \frac{1}{\beta} \nabla_x f_w \cdot J_f F_w J_f^t\,\nabla_x f_w
\end{equation}
where $\nabla_x f_w(x)$ is the Jacobian of the representation given the input, and $J_f(x)$ is the Jacobian of the representation with respect to the weights.
Decreasing the Fisher Information that the weights contain about the training set decreases the effective information between inputs and activations. However, the vice-versa may not be true in general. In fact, it is sufficient that $\norm{ \nabla_x f_w }$ decreases. Indeed, this fact was used in several works to enhance model robustness~\cite{virmaux2018lipschitz, fazlyab2019efficient}. However, as we show in~\Cref{fig:fisher_cifar}, AT reduces information in the features as the embedding defined by $|F_w^{-1}|$, that is, the log-variance of parameters is increased when increasing the $\varepsilon$ applied on training. Experiments are done with a ResNet-18 on CIFAR-10.
Interestingly, this provides the evidence that it is possible to achieve robustness without reducing $\norm{ \nabla_x f_w }$.

\section{Does invertibility contradict IB?}
\label{sec:invertibility}
\mt{Robust representations are (almost) invertible, even for out-of-distribution data~\cite{engstrom2019learning}.
~\Cref{fig:inv_all_model} shows examples of inversions using~\Cref{eq:inversion}. However, past literature claims that a classifier should store only information useful for the task. This is even more surprising as robust features should discard useful details more than standard models. 
This fact empirically proves that it is not necessary to remove information about the input to generalize well~\cite{invnet}. 
Moreover, when $f$ is an invertible map, the Shannon information $I(x,z)$ is infinite.
So, how can invertibility and minimality of representations be conciliated? Where is the excess of information which explains the gap?
As shown in~\Cref{sec:at_reduce}, the main problem of standard IB, is that it requires to operate in the activations during training and there is no guarantee that information is also reduced at test time, which is not as AT shows. 
The crucial point shown in~\cite{achille2019information} and in the previous sections, is that it is still possible to maintain information about input at test time while making the information inaccessible for the classifier. Moreover, an important result in this paper, is that it is possible to visualize the images that are effectively "seen" by the classifier in computing the prediction. By leveraging~\Cref{def:eff_info}, we define the \emph{effective image}.

\begin{definition}[Effective image]
\label{def:eff_images}
Let $\bar{z}=f_w(x)$, and let $f_w$ be the model trained with $\norm{\delta}_2 \leq \varepsilon$. 
We define effective image $x_{\text{eff},\varepsilon}$ at level $\varepsilon$, the solution of the following problem:
\begin{equation}
\label{eq:effective_image}
x_{\text{eff},\varepsilon} (x; z) = \argmin_{x'}{\norm{f_{w+n}(x) - f_w(x')}_2} 
\end{equation}
where $n \sim N(w, \Sigma^\star)$ and $\Sigma^\star = \beta F^{-1}(w)$. 
\end{definition}
The idea under effective images is to simulate the training conditions by artificially injecting the noise that approximates SGD. In this manner we can visualize  how  AT controls the conveyed information.
In~\Cref{fig:inv_all_model} we show some examples. Interestingly, robust features are not always good features: in fact, due to the poor diversity of the dataset (CIFAR-10), the feature \textit{color green} is highly correlated with class \textit{frog}.} 

\begin{figure}[htbp]
    \centering
    \includegraphics[width=0.38\textwidth]{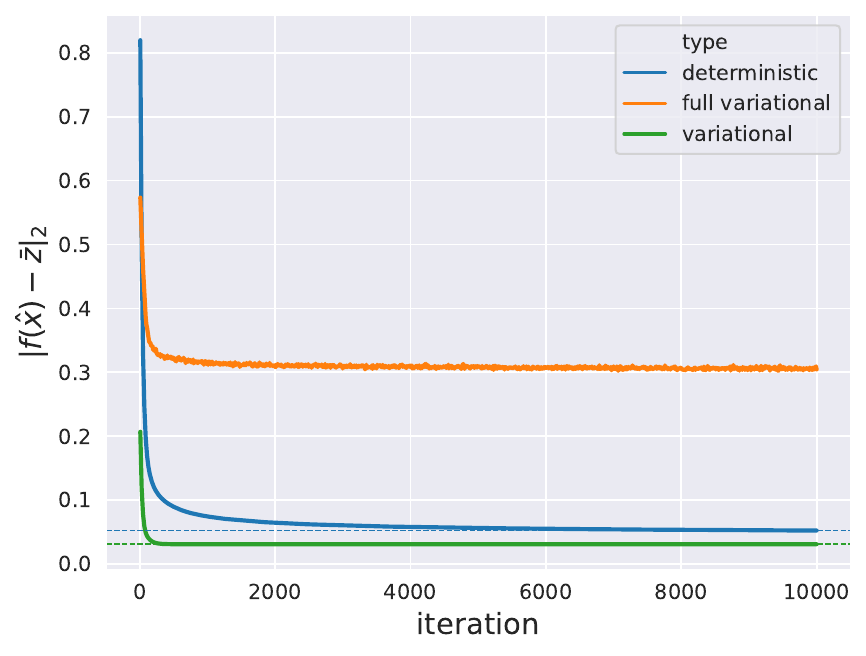}
    \caption{Comparison of $L_{inv} (x, z)$ of (orange) Effective images,  (green) variational and (blue) deterministic models.}
    \label{fig:loss_inv}
    \vspace{-10pt}
\end{figure}

\vspace{-3mm}
\paragraph{Adding effective noise (once) improves inversion}
\label{sub:once}
The quality of inversion depends on the capability of gradient flow to reach the target representation $\hat{z}$. Starting from regions that are distant from training and test points $f_w$ may be less smooth. Intuitively, especially during the first phase of optimization, it can be beneficial to inject noise to escape from local minima. Surprisingly, we discover that by injecting effective noise once, reconstruction is much faster and the quality of images improves dramatically. At the beginning of optimization, we perturb weights with $\bar{n} \sim N(0, \Sigma^\star)$ and solve the inversion with $f_{w+\bar{n}}$. By visually comparing row 2 and 3 of ~\Cref{fig:inv_all_model}, it is easy to see that injecting noise as described above, improves the quality of reconstruction. In support of this, in~\Cref{fig:loss_inv} we numerically assess the quality of representations using the loss $L_{inv} (x, z)$. 
The variational model, besides improving quality of fit, also allows fast convergence: convergence is achieved after roughly 200 iterations while the deterministic model converges after 8k iterations ($\sim 40\times$).

\begin{table*}[t]
\centering
\begin{tabular}{lrrrr}
\toprule
\multicolumn{5}{c}{\textbf{\ct}} \\
  &  \co &  \fmn &  \mn &  \sv        \\
 \midrule
 \textbf{Rob} &     \bff{44.92} &   \bff{76.89} &  \bff{88.11} & \bff{58.34} \\
  \textbf{St} &     35.76 &   67.15 &  64.17 & 36.6\\
\bottomrule
\end{tabular}
\begin{tabular}{rrrr}
\toprule
\multicolumn{4}{c}{\textbf{\co}} \\
\ct &  \fmn &  \mn &  \sv        \\
\midrule
 74.47 &  \tbf{84.85} &  \tbf{94.96} & \tbf{70.61} \\
 \tbf{80.18} &   76.10 &  79.46 & 55.6 \\
\bottomrule
\end{tabular}
\caption{Transfer accuracy [$\%$] starting from CIFAR-10 (left) and CIFAR-100 (right).}
 \label{tab:transfer10}
\end{table*}
\begin{table*}[t]
\centering
\resizebox{1.0\textwidth}{!}{%
\begin{tabular}{lrrrrrrrrrrr}
\toprule
\textbf{\im} &   \ct &  \co &  \fmn &  \mn &  \sv  & \air &  \bir &  \ca &  \dog &  \fl &  \ind \\
\midrule
 \rob &  \bff{93.78} &     \bff{77.94} &   \bff{90.09} &  \bff{98.03} & \bff{76.90} & 33.81 &          35.91 & 40.47 & 66.25 &    93.15 &   63.06 \\
  \sta &  84.72 &     64.48 &   86.38 &  93.91 & 50.46 & \bff{36.72} & \bff{53.58} & \bff{50.12} & \bff{88.39} &    \bff{95.48} &   \bff{72.84} \\
  \bottomrule
\end{tabular}
}

\caption{Transfer accuracy [$\%$] of a ResNet50 pretrained on ImageNet.}
 \label{tab:transferImage}
\end{table*}
\vspace{-2mm}
\begin{table*}[htbp]
\centering
\begin{tabular}{llrrrr}
\toprule
\multicolumn{6}{c}{\textbf{ResNet50}} \\
 & \tbf{\co} &\ct &  \fmn &  \mn &  \sv        \\
\midrule
0 & \rob &  74.47 &   \tbf{84.85} &  \tbf{94.96} & \tbf{70.61} \\
  & \sta &  \tbf{80.18} &   76.10 &  79.46 & 55.60 \\ \midrule
1 & \rob &    85.67 &   \tbf{89.22} &  \tbf{98.33} & \tbf{91.34} \\
  & \sta &  \tbf{87.80} &   88.65 &  97.75 & 91.12 \\ \midrule
2 & \rob &    94.82 &   \tbf{92.58} &  \tbf{99.24} & \tbf{96.63} \\
  & \sta &  \tbf{95.20} &   91.78 &  99.22 & 96.60 \\
\bottomrule
\end{tabular}
\begin{tabular}{rrrr}
\toprule
\multicolumn{4}{c}{\textbf{ResNet18}} \\
\ct &  \fmn &  \mn &  \sv        \\
\midrule
68.89 &   \tbf{83.40} &  \tbf{94.61} & \tbf{61.08} \\
\tbf{76.50} &   76.30 &  77.98 & 49.32 \\
 \cmidrule{1-4}
82.40 &   \tbf{87.59} &  97.82 & \tbf{89.68} \\
\tbf{85.11} &   86.11 &  \tbf{97.84} & 88.62 \\
 \cmidrule{1-4}
94.59 &   \tbf{92.48} &  \tbf{99.30} & \tbf{96.39} \\
\tbf{95.10} &   92.03 &  99.15 & 96.29 \\
\bottomrule
\end{tabular}
\caption{Performance comparison using different architectures transfering from CIFAR-100.}
\label{tab:arch_diff_main}
\end{table*}

\section{Transferability-accuracy trade off}
\label{sec:transfer}
The insights from the previous sections motivate the following argument: if in robust models information is still there, is it possible that features not useful for the original task $y_1$ are useful for other tasks? 
In a sense, $z$ is a well-organized semantic compression of $x$ such that it approximately allows to linearly solve the new task $y_2|z$. 
How well the task $y_2$ is solved depends on how $z$ is organized. 
In fact, even though $z$ is optimal for $y_1$ and for reconstructing $x$, it still could be not optimal for $y_2$. This intuition suggests that having robust features $z$ is more beneficial than having a standard model when the distance $d(y_2, y_1)$ between tasks $y_2$ and $y_1$ is such that features from the source models are not easily adaptable to the new task.
Thus, there may exist a trade-off between accuracy on a given task and stability to distributions changes: "locally", standard models work better as feature extractor, but globally this may not be true. 
\mt{In order to test our hypothesis, we (i) analyze the structure of representations extracted from adversarially-trained models, (ii) provide a theoretical motivation and (iii) experimentally confirm the theory by showing the emergence of a trade-off in transferability.}

\begin{figure}[!htbp]
\vspace{-4mm}
    \centering
    \begin{subfigure}[t]{0.23\textwidth}
    \includegraphics[width=\textwidth]{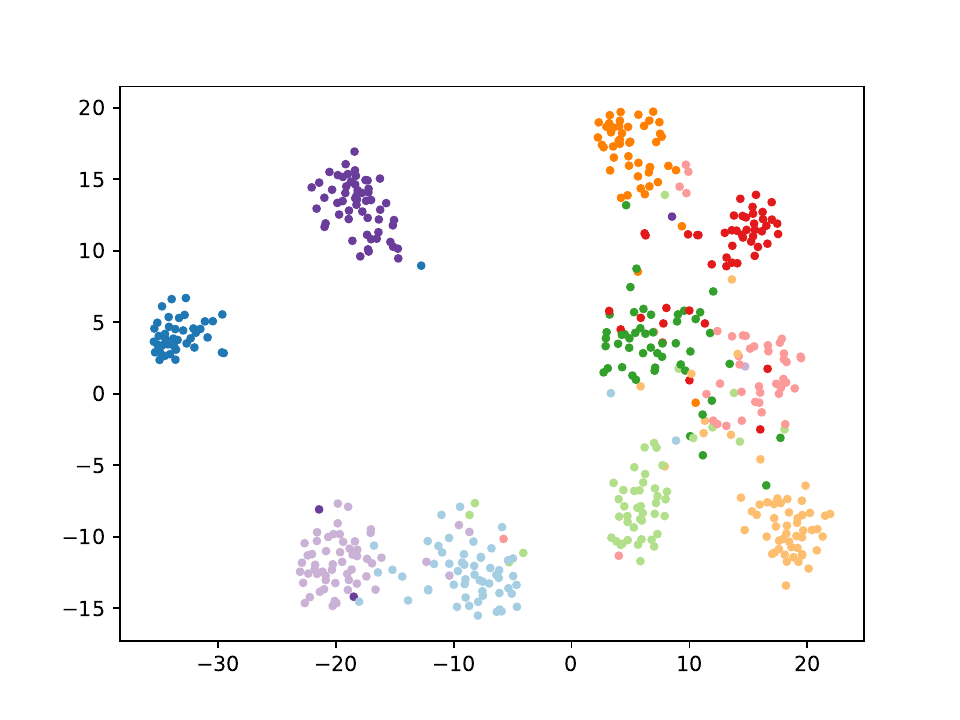}
    \caption{$\varepsilon=0.5$}
    \end{subfigure}
    \hspace*{-6mm}
    \begin{subfigure}[t]{0.23\textwidth}
    \includegraphics[width=\textwidth]{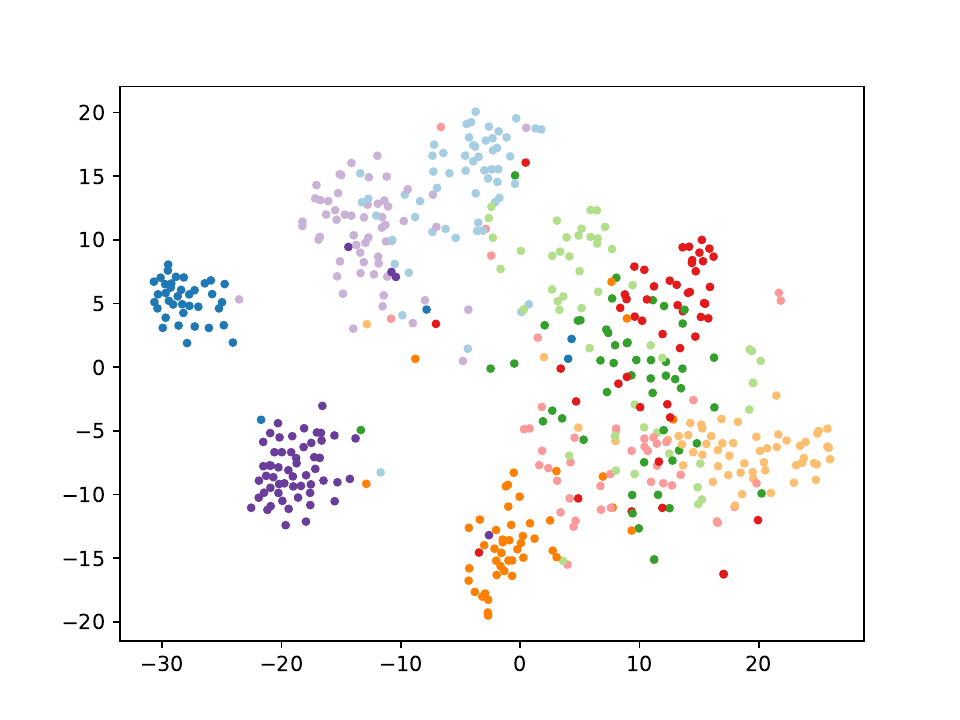}
    \caption{$\varepsilon=1$}
    \end{subfigure}
    \begin{subfigure}[t]{0.23\textwidth}
    \includegraphics[width=\textwidth]{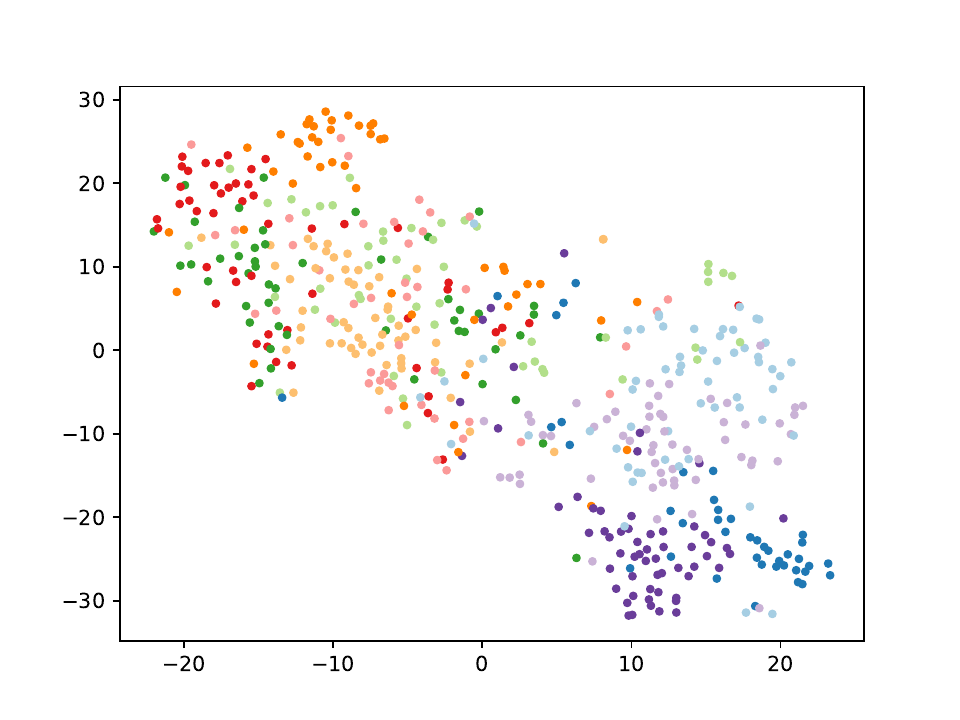}
    \caption{$\varepsilon=2$}
    \end{subfigure}
    \hspace*{-6mm}
    \begin{subfigure}[t]{0.23\textwidth}
    \includegraphics[width=\textwidth]{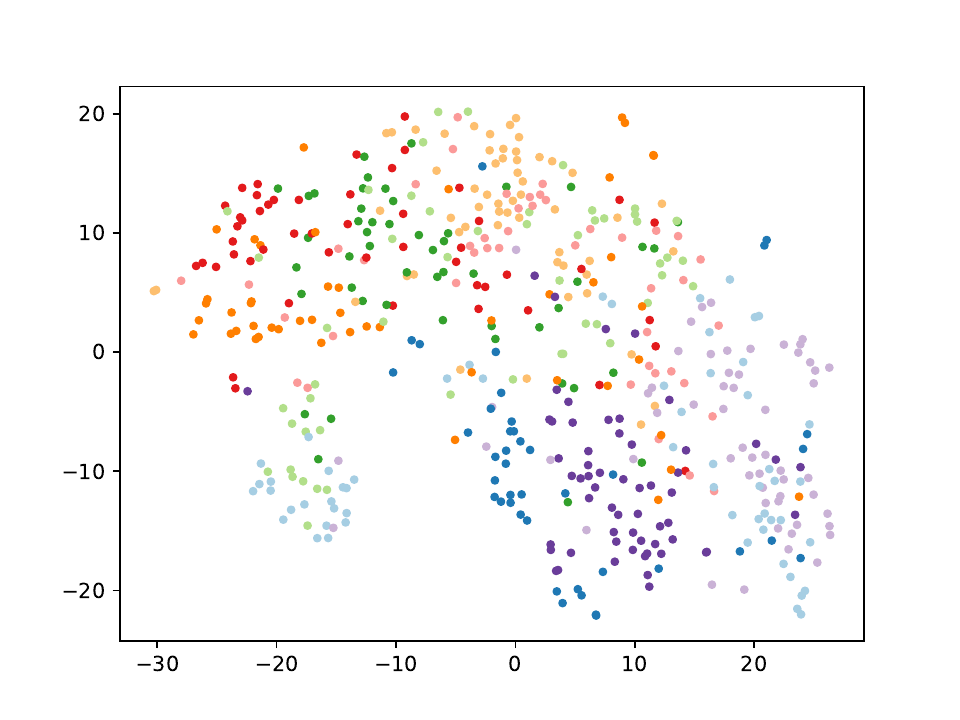}
    \caption{$\varepsilon=3$}
    \end{subfigure}
    \caption{t-SNE of features extracted from a batch of 512 images with a robust ResNet-18 model trained on CIFAR-10 for different values of $\varepsilon$. As $\varepsilon$ increases, features become less discriminative.
    }
    \label{fig:tsne}
\end{figure}

\noindent Recently,~\cite{frosst2019analyzing} showed that more entangled features, that is more class-independent, allow for better generalization and robustness.
In order to understand the effect of AT,~in \Cref{fig:tsne} we show the t-SNE~\cite{maaten2008visualizing} embedding of final representations for different values of $\varepsilon$: as $\varepsilon$ increases, the entanglement increases at the expenses of less discriminative features. Thus, robust models capture more high-level features instead of the ones useful only for the task at hand.

\newpage
\subsection{Effective transferable information}
\mt{Interestingly, Fisher Information theory presented in~\Cref{sec:at_reduce} can be applied even to provide an theoretical intuition about transferability of robust models.}

Since AT reduces $F_{w\mid D}$, it reduces the information that the network has about the dataset $\D$. In fact:
 \begin{equation}
    \label{eq:shannon-fisher-w-y}
        I(w;\D) \approx H(w) - \E_\D\Big[\frac{1}{2} \log\Big(\frac{(2\pi e)^k}{|F_{w|\D}|}\Big)\Big], 
\end{equation}
where $F_{ w|\D } \approx \nabla_\D w^t F_w \nabla_\D w$.
From the previous proposition we can see that there are two ways of reducing the information $I(w;\D)$. 
The first is reducing $|F_w|$ and the other is making the weights $w$ more stable with respect to perturbation of the datasets. For example, the latter can be accomplished by choosing a suitable optimization algorithm or a particular architecture.
Reducing the Fisher $F_{w|D}$, implies that the representations vary less when perturbing the dataset with $\delta \D$. 
This explains that fact that AT is more robust to distribution shifts. 
We would like to remark again that there are two ways for transferring better: one is to reduce $\norm{\nabla_\D w}_2$ and the other one is reducing $|F_w|$. 

\subsection{Transferability experiments}
\begin{figure}
\centering
    \includegraphics[width=0.75\columnwidth]{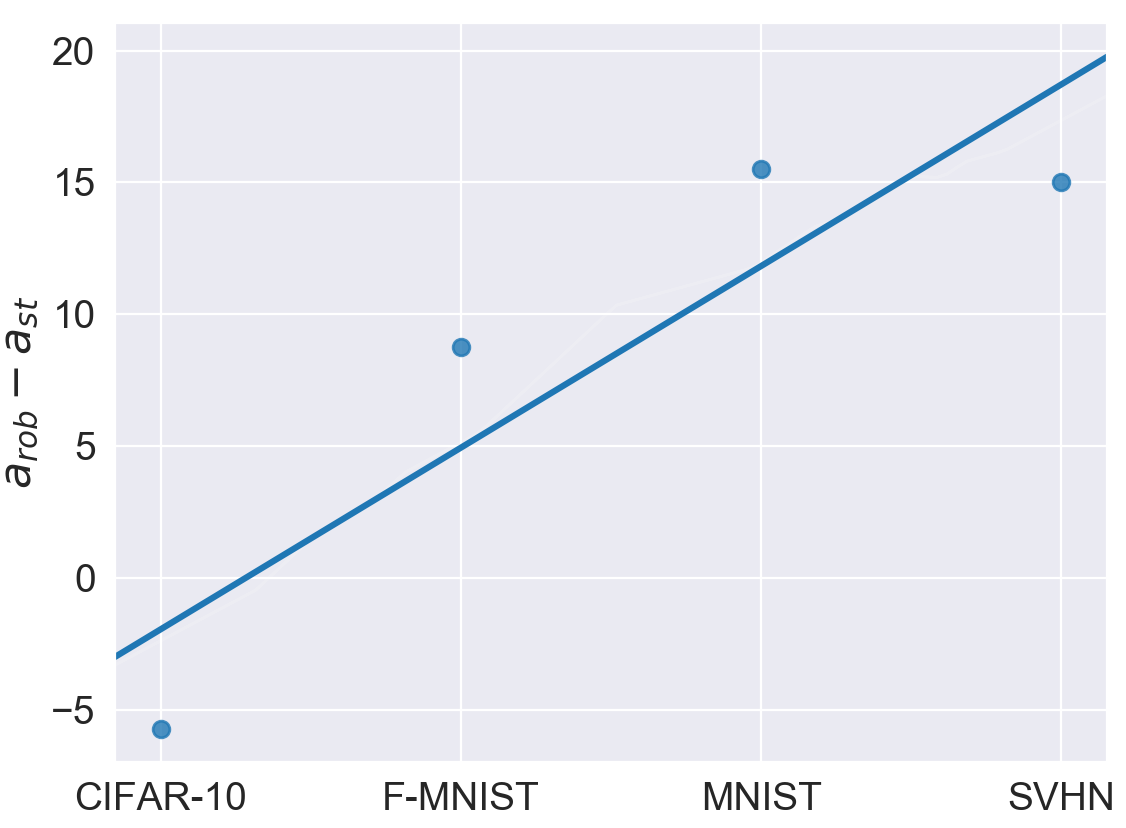}
  \caption{Accuracy gap between the robust and standard model transfering from CIFAR-100.}
  \label{fig:c100_gap}         
 \end{figure}

We employ CIFAR-10~\cite{cifar}, CIFAR-100~\cite{cifar} and ImageNet~\cite{deng2009imagenet} as source datasets. All the experiments are obtained with ResNet-50 and $\varepsilon=1$ for CIFAR and $\varepsilon=3$ for ImageNet as described in~\cite{ilyas2019adversarial} and in the Appendix. In \Cref{tab:transfer10} we show performance of fine-tuning for the networks pretrained on CIFAR-10 and CIFAR-100 transferring to CIFAR-10, CIFAR-100 F-MNIST~\cite{xiao2017fashion}, MNIST~\cite{mnist} and SVHN~\cite{svhn}.
Details of target datasets are given in Appendix.
Results confirm our hypothesis: when a task is "visually" distant from the source dataset, the robust model performs better. For example, CIFAR-10 images are remarkably different from the SVHN or MNIST ones.
Moreover, as we should expect, the accuracy gap (and thus the distance) is {\em not} symmetric: while CIFAR-100 is a good proxy for CIFAR-10, the opposite is not true. 
In fact, when fine-tuning on a more complex dataset, from a robust model is possible to leverage features that the standard model would discard.
According to~\cite{cui2018large}, we employ Earth Mover’s Distance (EMD) as a proxy of dataset distance, and we extract the {\em order} between datasets. As we show in \Cref{fig:c100_gap}, the distance correlates well with the accuracy gap between robust and standard {\em across} all the tasks. 
\Cref{tab:transferImage} shows similar results using models pretrained on ImageNet. The robust model provides better performance in all the benchmarks being them quite different from the original tasks. We also report experiments  on more difficult datasets namely Aircraft \cite{aircrafts}, Birds \cite{birds}, Cars \cite{cars}, Dogs~\cite{dogs}\footnote{The Stanford Dogs has been built using images and annotations from ImageNet.}, Flowers \cite{flowers}, Indoor \cite{indoor67} that would have not been suitable for transfering from simpler tasks like CIFAR-10 and CIFAR-100. 
Not surprisingly the robust model shows lower accuracy compared to the standard one since images are very similar to those contained in the ImageNet dataset. For examples, Dogs images are selected from ImageNet. Also with ImageNet, as shown by~\Cref{fig:img_acc_order}, the difference in accuracy between the two model is correlated with distance.
We can see that the furthest the task the higher the difference in accuracy in favor of the robust model.
For the sake of space, we report similar results for other source and target datasets in the Appendix. Finally, in~\cref{tab:arch_diff_main} we analyze the impact of using a bigger architecture.  It is noticeable that with the more complex network (ResNet50) the gap is reduced in  cases  where  the  standard  model  is  better  and  it  is increased in cases where the robust one is better. 

\paragraph{Robustness of fine-tuned models} Are the fine-tuned models still robust? As already experimentally shown by~\cite{ilyas2019adversarial,shafahi2019adversarially}, an advantage of using $f_w(\cdot)$ as a feature extraction is that then the new model $A_2 f_w(\cdot) + b_2$ is robust for the new task. Indeed, it is sufficient to show that the Fisher $F_{y|x}$ is bounded from above by $F_{z|x}$, that is, the linear classifier can only reduce information.
\begin{lemma}
\label{lemma:robustness}
Let $z = f_w$ be the feature extractor, $y = Az + b$, with $A \in \mathbb{R}^{k \times p}$, where $k < p$. Let $F_{z|x}$ be the Fisher of its activations about the input. Then, it holds:
$
\tr{F_{y|x}} \le \tr{F_{z|x}}
$.
\end{lemma}

\section{Conclusions}
Existing works about robust models~\cite{madry2017towards,ilyas2019adversarial,tsipras2018robustness} showed that there exists a trade-off between robustness of representations and accuracy for the task. This paper extends this property showing the parameters of robust models are the solution of a trade-off between usability of features for other tasks and accuracy for the source task. By leveraging results in~\cite{achille2019information,achille2018emergence}, we show that AT has a compression effect similarly to IB, and we explain how a network can be invertible and lose accuracy for the task at the same time. Moreover, we show that AT also reduces information in the weights, extending the notion of effective information from perturbations of the weights, to perturbations of the input. 

We also show that effective noise can be also useful to improve reconstruction of images both in terms of convergence and quality of reconstruction.

Finally, we provide an analytic argument which explains why robust models can be better at transferring features to other tasks. As a corollary of our analysis, to train a generic feature extractor for several tasks, it is best to train adversarially, unless one already knows the specific task for which the features are going to be used.


\pagebreak

\section*{Acknowledgments}
We would like to thank Stefano Soatto for providing valuable feedback on the manuscript.
\bibliography{main}

\bigskip
\clearpage
\appendix 

\section{Complementary propositions}

\begin{proposition}[Fisher in the weights, \cite{achille2019information}]
  \label{prop:fisher-complexity-equivalence}
	Let $P(w) \sim N(0, \lambda^2 I)$ and $Q(w|\D) \sim N(w^*, \Sigma)$, centered at any local minimizer $w^*$ of the cross-entropy loss obtained with any optimization algorithm.
	For a sufficiently small $\beta$, the covariance $\Sigma$ that minimizes $C_\beta(\D;P,Q)$ is
		\[
		   \textstyle \Sigma^* = \frac{\beta}{2} \left(H + \frac{\beta}{2\lambda^2}I \right)^{-1}.
		\]

		For $\Sigma=\Sigma^*$, the Information in the Weights is
		\begin{align}
	    \label{eq:kl-equal-fisher}
		&\KL{Q}{P} = \frac{1}{2} \log \left|H + \frac{\beta}{2\lambda^2}\right| + \frac{1}{2} k \log \frac{2\lambda^2}{\beta} -k \nonumber \\
		 & \quad + \frac{1}{2\lambda^2}\left[ \|w^*\|^2 + \tr\left[ \frac{\beta}{2} \left( H + \frac{\beta}{2\lambda^2}I \right)^{-1} \right] \right].
		\end{align}

		If most training samples are predicted correctly, we can estimate \cref{eq:kl-equal-fisher} by substituting $H \approx F$.
\end{proposition}

\begin{proposition}[\cite{achille2019information}]
\label{prop:re-emergence}
For small values of $\beta$ we have:
\begin{enumerate}[(i)]
    \item The Fisher Information $F_{z|x} = \E_z [\nabla^2_x \log p(z|x)]$ of the activations w.r.t. the inputs is:
    \[F_{z|x} = \frac{1}{\beta} \nabla_x f_w \cdot J_f F_w J_f^t\,\nabla_x f_w,\]
    where $\nabla_x f_w(x)$ is the Jacobian of the representation given the input, and $J_f(x)$ is the Jacobian of the representation with respect to the weights. In particular, the Fisher of the activations goes to zero when the Fisher of the weights $F_w$ goes to zero.
    \item Under the hypothesis that, for any representation $z$, the distribution $p(x|z)$ of inputs that could generate it concentrates around its maximum, we have:
    \begin{equation}
        I_{\text{eff},\beta}(x;z) \approx H(x) - \E_x\Big[\frac{1}{2} \log\Big(\frac{(2\pi e)^k}{|F_{z|x}|}\Big)\Big],
    \end{equation}
    hence, by the previous point, when the Fisher Information of the weights decreases, the effective mutual information between inputs and activations also decreases.
\end{enumerate}
\end{proposition}

\section{Proofs of propositions}
In the following we prove~\Cref{lemma:robustness}.
\begin{proof}{\textbf{Proof of~\Cref{lemma:robustness}}} 
\label{proof:robustness}
The intuition under this lemma is very similar to Data Processing Inequality (DPI). If we have $X \to Z \to Y$ and $Z$ is robust, the map $Y\to Z$ can only use robust information. As shown, for example, in~\cite{zegers2015fisher} [Theorem 13], the DPI also holds for the Fisher Information. 
\end{proof}
Although the previous lemma is very simple, it has remarkable consequences: as soon as one is able to extract robust features, at some level of the "chain", then all the information extracted from these features is robust. For example,~\cite{ilyas2019adversarial} shows that by training on images that are obtained by robust models, leads to a robust model, without applying AT. In this case, the robust features are directly the images.



\section{Experimental setting}
To quantitatively evaluate the improved  transferability provided by robust models we perform experiments on common benchmarks for object recognition. More in details, we fine tune three networks pretrained on CIFAR-10, CIFAR-100 and ImageNet. 

We used the pretrained robust ResNet-50 models on CIFAR-10 (with $\varepsilon=1$ and ImageNet (with $\varepsilon=3$) from~\cite{ilyas2019adversarial}.
Similarly, we trained on CIFAR-100 with $8$ steps of PGD iterations with $\varepsilon=1$.

We fine-tune with different modalities: 0) both the linear classifier and the batch norm before it, 1) both the linear classifier and the batch norm of the entire network, 2) the entire network. We then compare the top1 accuracy on the test set of the different models. 
We asses the performance on the tranferability using a Resnet50. 
For CIFAR 10 and CIFAR 100 fine tuning is done for 120 epochs using SGD with batch size 128, learning rate that starts from 1e-2 and drops to 1e-3, 1e-4 at epochs 50 and 80 respectively. We use weight decay 5e-4.
For Imagenet fine tuning is done for 300 epochs with batch size equal to 256, the same learning rate decay at epochs 150 and 250 respectively and weight decay 1e-4. 
We use momentum acceleration with parameter 0.9 for all datasets. \\
In~\Cref{tab:datasets} we report the description of the datasets used in this paper.
\begin{table*}[h]
    \centering
    \begin{tabular}{llrrr}
    \toprule
         Dataset&Task Category&Classes&Training size& Test size  \\
         \midrule
         Imagenet \cite{deng2009imagenet} & general object detection &1000& 1281167 & 50000 \\
         CIFAR-10 \cite{cifar} &general object detection& 10&50000&10000 \\
         CIFAR-100 \cite{cifar} &general object detectio&100&50000&10000 \\
         MNIST \cite{mnist}&handwritten digit recognition&10&60000&10000 \\
         F-MNIST \cite{xiao2017fashion}&clothes classification&10&60000&10000 \\
         SVHN \cite{svhn} &civic number classification&10&73257&26032 \\ \hline
         Oxford Flowers~\citep{flowers}      & fine-grained object recognition  & 102       & 2,040         & 6,149   \\
         CUB-Birds 200-2011~\citep{birds}        & fine-grained object recognition  & 200       & 5,994         & 5,794   \\
         FGVC Aircrafts~\citep{aircrafts}      & fine-grained object recognition  & 100       & 6,667         & 3,333   \\
         Stanford Cars~\citep{cars}       & fine-grained object recognition  & 196       & 8,144         & 8,041   \\
         Stanford Dogs~\citep{dogs}       & fine-grained object recognition  & 120       & 12,000        & 8,580   \\
         MIT Indoor-67~\citep{indoor67}       & scene classification             & 67        & 5,360         & 1,340   \\
         \bottomrule
    \end{tabular}
    \caption{Datasets employed in this paper.}
    \label{tab:datasets}
\end{table*}

\section{Image reconstruction}
\subsection{Algorithms}
~\Cref{alg:eff_images} shows the procedure to compute effective images (see~\Cref{def:eff_images}), while~\Cref{alg:var_inv} represents the procedure to compute the variational inversion where noise in sample once.

\begin{algorithm}
\begin{algorithmic}[H!]
    \STATE {\bfseries Input:} Image $x$, representation $\bar{z} = f_w(x)$, noise matrix $\Sigma^\star = \beta F^{-1}(w)$, number of steps $N$, learning rate $\eta$.
    \STATE Initialize $r_0 \sim N(0.5, \sigma)$
    \STATE Sample $\bar{n}\sim N(0,\Sigma^\star)$
    \FOR{each iteration $k < N$}
    \STATE Reconstruct the image at step $k$ through SGD on input input space:
    $$
    r_{k+1} = r_k - \eta \nabla_x \norm{f_{w+\bar{n}}(r_k) - \bar{z}}_2
    $$
    \ENDFOR
    \STATE Return $\hat{x} = r_N$
\end{algorithmic}
\caption{"Variational" Inversion. Injecting noise once before the optimization, improves the image reconstruction.}
\label{alg:eff_images}
\end{algorithm}
\begin{algorithm}
\begin{algorithmic}[H!]
    \STATE {\bfseries Input:} Image $x$, representation $\bar{z} = f_w(x)$, noise matrix $\Sigma^\star = \beta F^{-1}(w)$, number of steps $N$, learning rate $\eta$.
    \STATE Initialize $r_0 \sim N(0.5, \sigma)$
    \FOR{each iteration $k < N$}
    \STATE {\bf 1.} Sample $n\sim N(0,\Sigma^\star)$
    \STATE {\bf 2.} Reconstruct the image at step $k$ through SGD on input input space:
    $$
    r_{k+1} = r_k - \eta \nabla_x \norm{f_{w+n}(r_k) - \bar{z}}_2
    $$
    \ENDFOR
    \STATE Return $\hat{x} = r_N$
\end{algorithmic}
\caption{Effective images. Inversion only capture statistics effectively used during training by the model.}
\label{alg:var_inv}
\end{algorithm}

\begin{figure}[htbp]
\centering
\includegraphics[width=0.9\columnwidth]{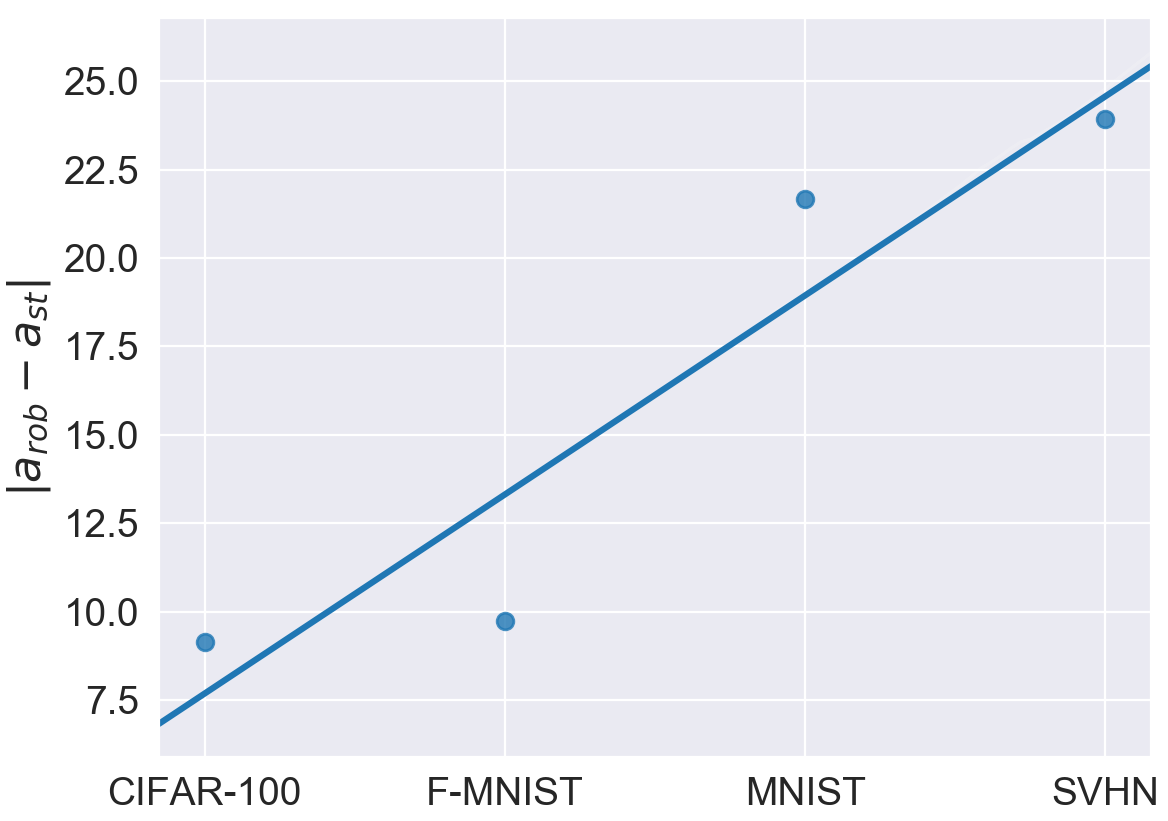}
\caption{Accuracy gap between the robust and standard model transfering from CIFAR-10.}
\label{fig:c_gap}       
\end{figure}

\subsection{Effect of $\varepsilon$ on the inversion}
\begin{figure*}[htbp]
    \centering
    \includegraphics[width=\textwidth]{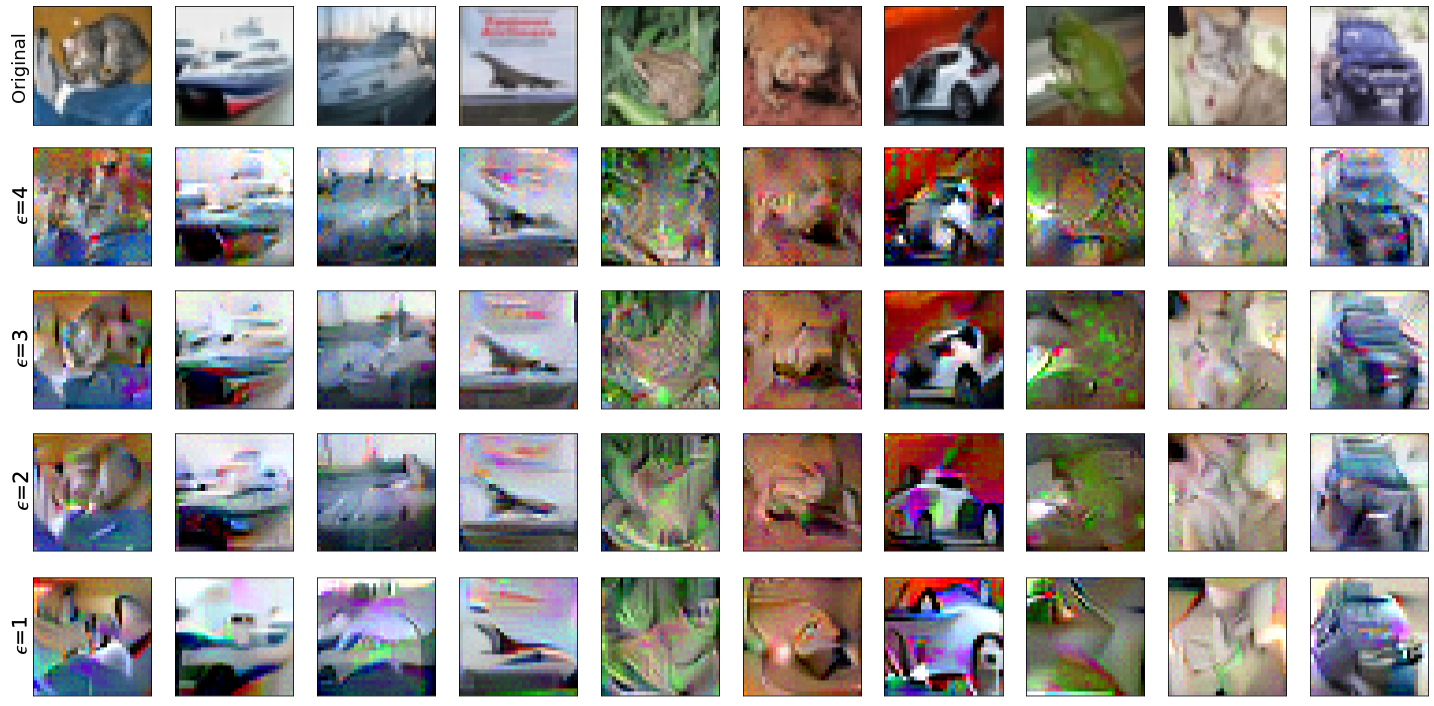} %
    \caption{Inversion of a ResNet-18 robust model with several values of $\varepsilon$. As we can see, there is an optimal value of $\varepsilon^\star$ for the reconstruction task. When $\varepsilon > \varepsilon^\star$ details start to be removed. Instead, when $\varepsilon < \varepsilon^\star$, the network is not regular enough to make the inversion feasible.}
    \label{fig:inv_vs_eps}
\end{figure*}

In~\Cref{fig:inv_vs_eps} it is shown the effect of training with different values of $\varepsilon$ on the image reconstruction.





\section{Omitted tables and figures}
We test the trivial hypothesis that standard models are better at transfering features when the source and target distributions are nearly the same: we choose CIFAR-10 as source dataset and CINIC-10~\cite{darlow2018cinic} as target dataset removing the images in common with CIFAR-10. The remaining images are extracted from ImageNet. We call this dataset CINIC-IMAGENET.
As~\cite{darlow2018cinic} shows, the pixel statistics are very similar, and in fact the standard models perform better at linear transfer:

\begin{table}[htbp]
    \centering
    \begin{tabular}{cc}
    \toprule
      \textbf{\sta}  & \textbf{\rob}\\
      \midrule
      84 & 80\\
    \end{tabular}
    \caption{ResNet-50 with CIFAR-10 as source dataset and CINIC-IMAGENET as target dataset. Accuracy in percentage after fine-tuning of linear layer.}
    \label{tab:cinic_vs_c10}
\end{table}


\section{Transfer with all modes}

\begin{figure*}[htbp]
\includegraphics[width=\textwidth]{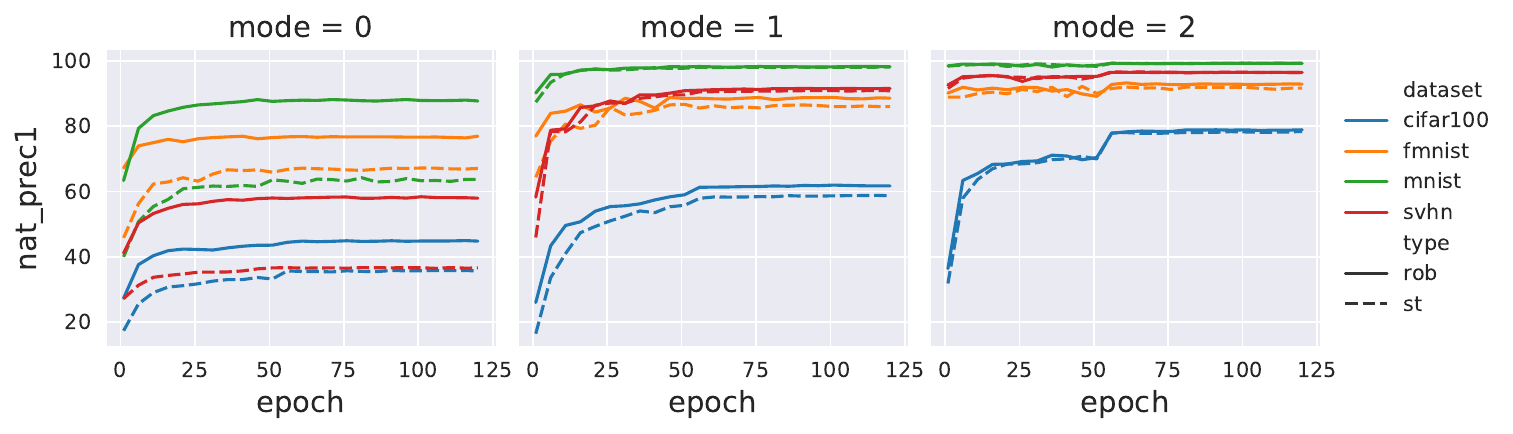}
\caption{Validation accuracy \% during fine tuning of a network pretrained on CIFAR-10.}
\label{fig:dyn_cifar10}       
\end{figure*}

\begin{figure*}[htbp]
\includegraphics[width=1\textwidth]{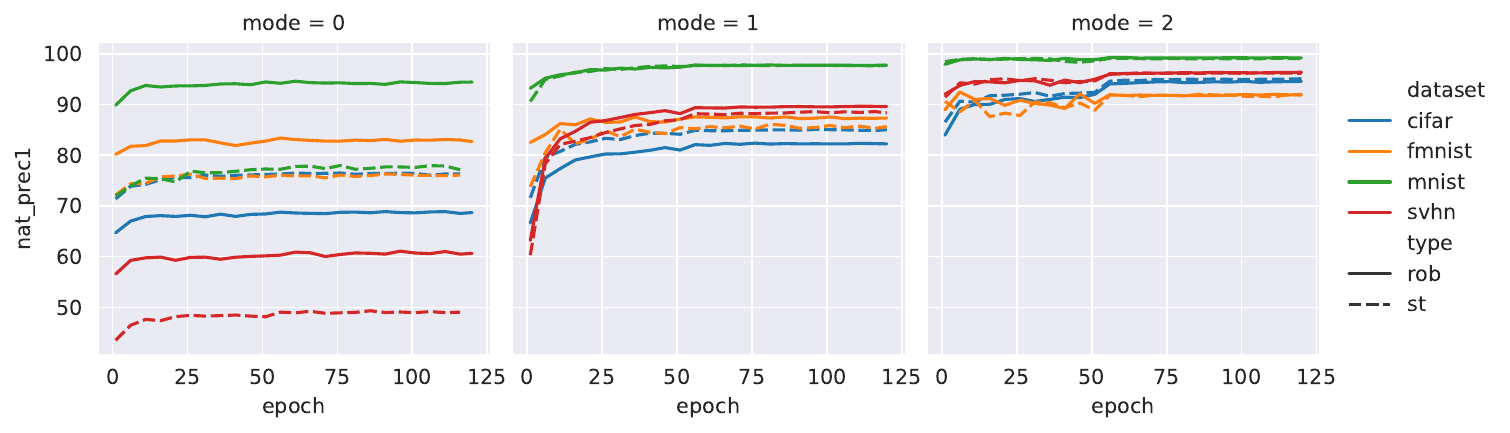}
\caption{Validation accuracy \% during fine tuning of a network pretrained on CIFAR-100.}
\label{fig:dyn_cifar100}       
\end{figure*}

\begin{figure*}[htbp]
\includegraphics[width=1.0\textwidth]{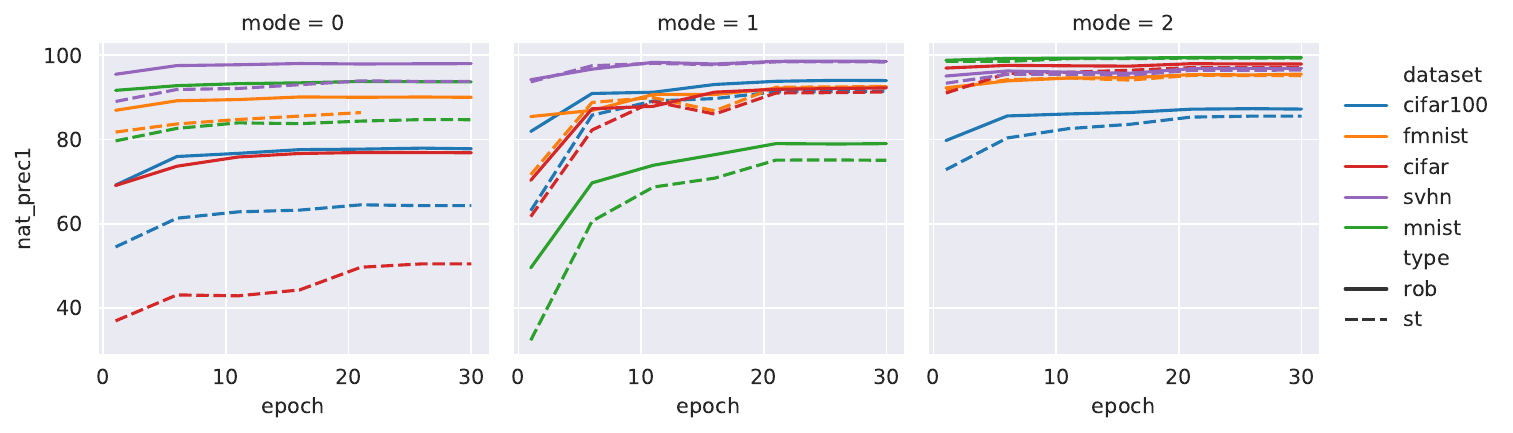}
\caption{Validation accuracy \% during fine tuning of a network pretrained on Imagenet.}
\label{fig:dyn_imagenet_simple}       
\end{figure*}

\begin{figure*}[htbp]
\includegraphics[width=1.0\textwidth]{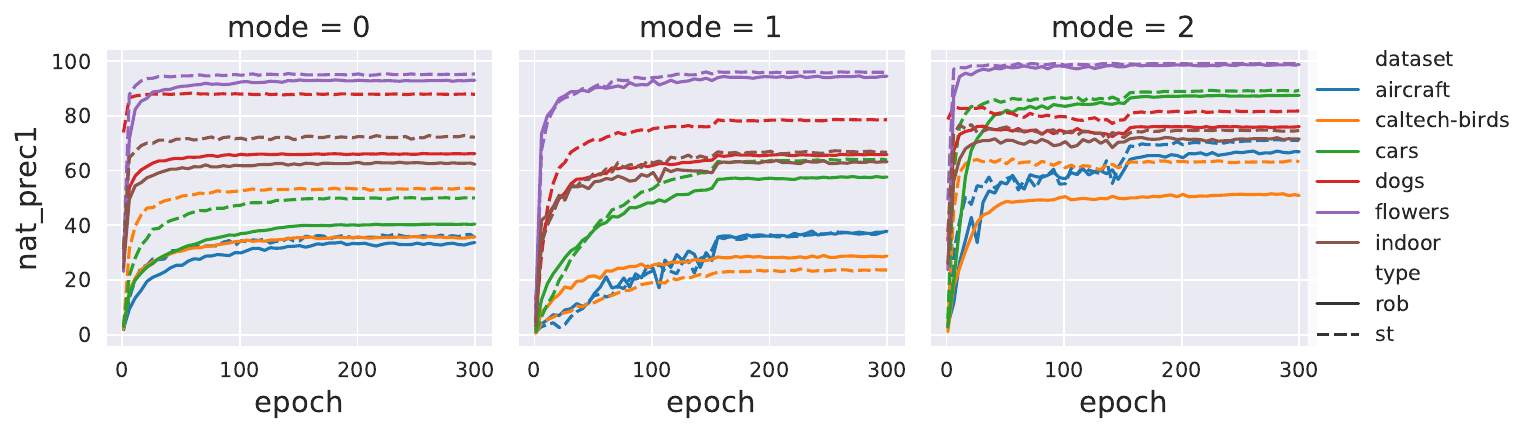}
\caption{Validation accuracy \% during fine tuning of a network pretrained on ImageNet.}
\label{fig:dyn_imagenet}       
\end{figure*}

\begin{figure*}[htbp]
 	 	\includegraphics[width=0.5\textwidth]{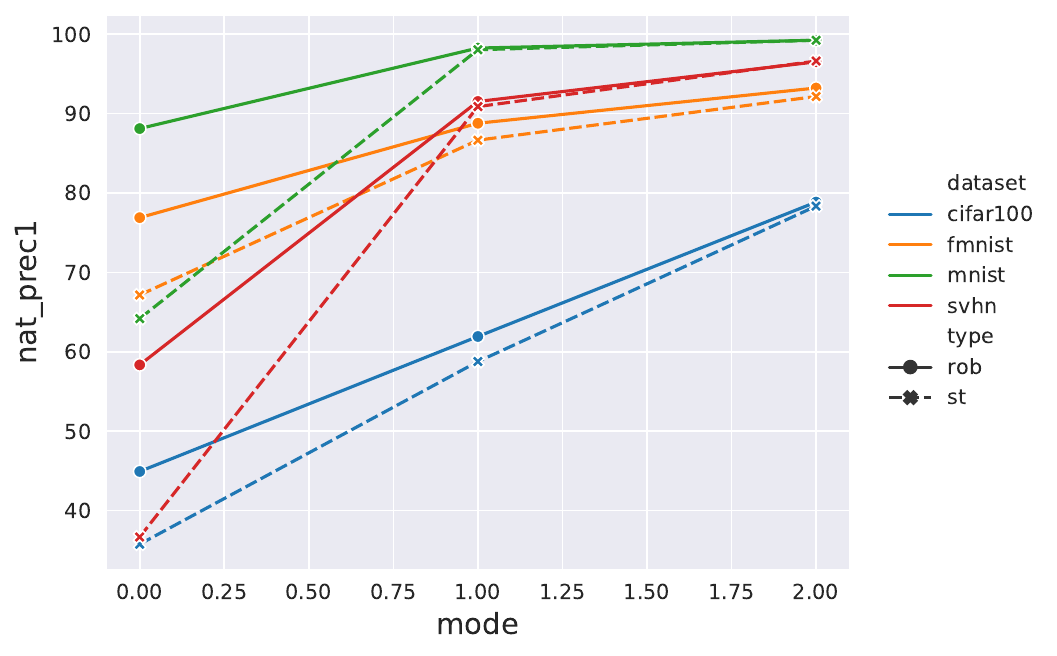}
 	 	\includegraphics[width=0.5\textwidth]{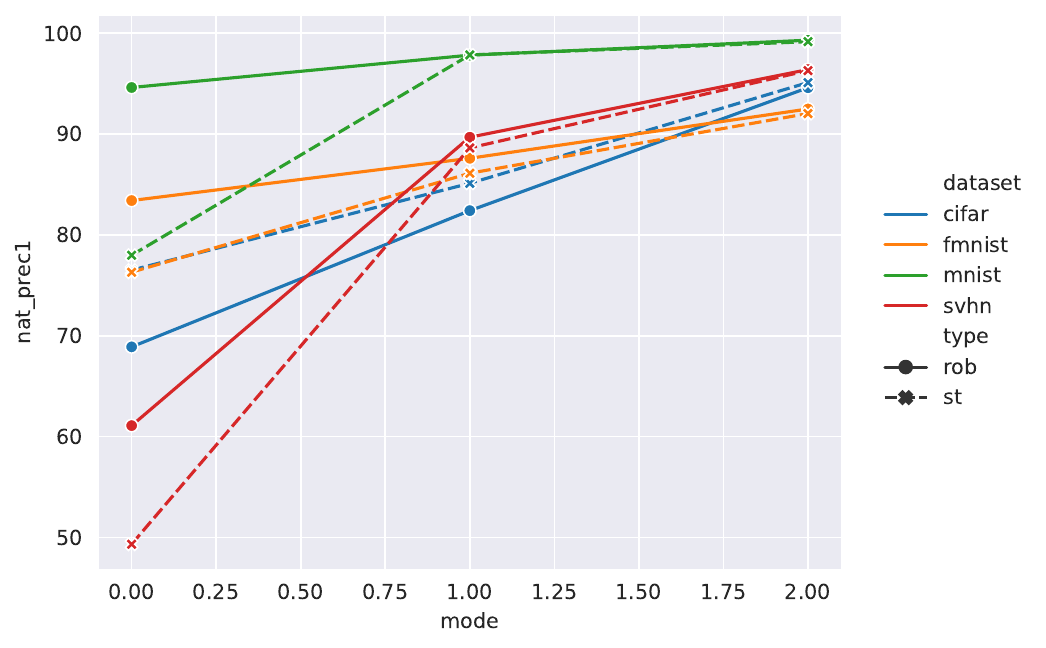}
  	 	\includegraphics[width=0.5\textwidth]{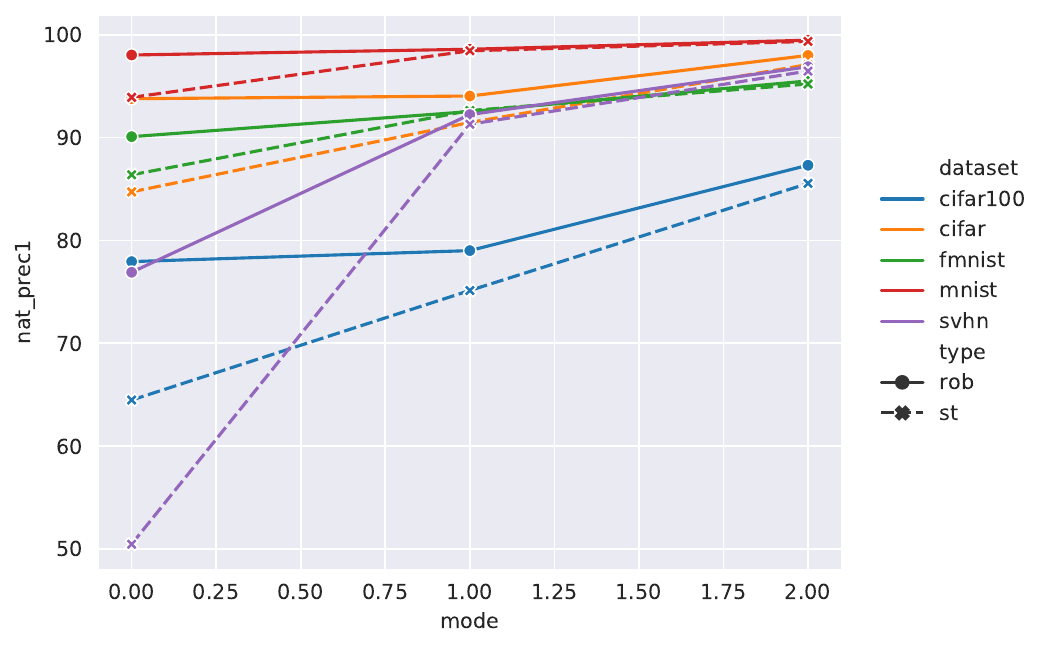}
 	 	\includegraphics[width=0.5\textwidth]{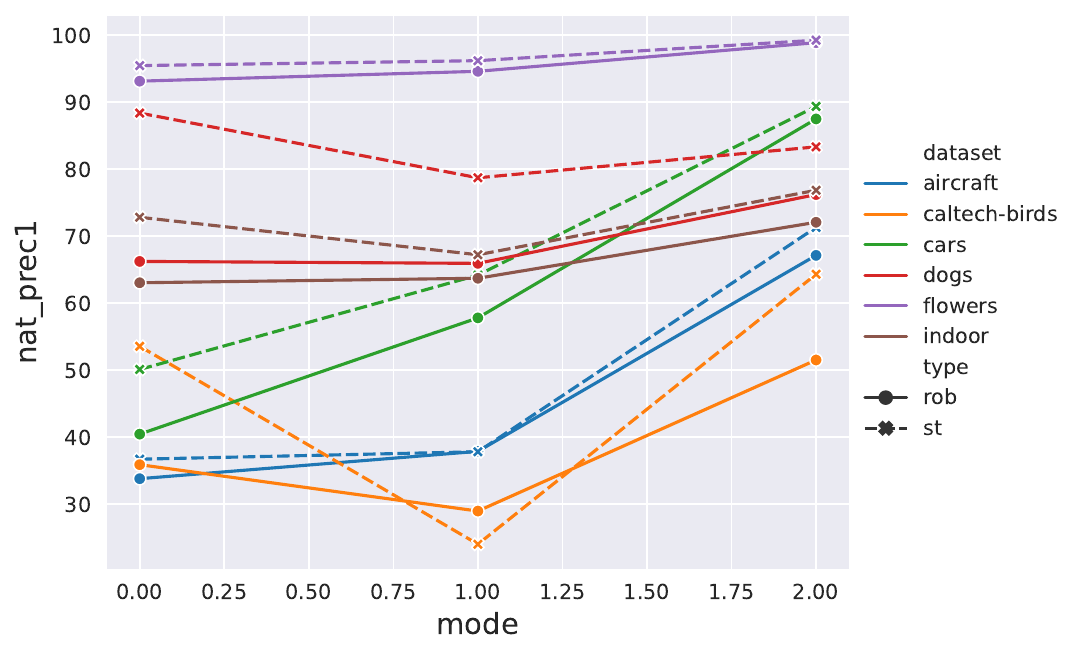}
\caption{Performance comparison in terms of accuracy \% using different fine tuning modes. First row: CIFAR-10 (left) CIFAR-100 (right); Second row: ImageNet.}
\label{fig:mode} 
\end{figure*}

While our aim is to show that robust models have better linear transferability than standard ones, we report here results also for fine tuning in modalities 1 and 2 (\Cref{tab:transfer10_ext,tab:transferImage_ext,tab:diffs,tab:transferImageSimplediff,tab:transferImagediff} and \Cref{fig:dyn_cifar10,fig:dyn_cifar100,fig:dyn_imagenet_simple,fig:dyn_imagenet,fig:mode}). Of course, the performance gap in these cases is reduced compared to mode 0 (see \Cref{tab:diffs,tab:transferImageSimplediff,tab:transferImagediff}) being the network able to change more to adapt to the new task. Interestingly, we notice a substantial impact of the batch norm layers on the classification performance: mode 1 provides a significant boost in classification accuracy compared to mode 0 particularly when the network is pretrained on simple datasets (CIFAR-10, CIFAR-100), even though the parameters of feature extractor are still kept fixed and only the batch norm in the entire network is fine tuned.
\begin{table*}[htbp]
\centering
\resizebox{0.95\textwidth}{!}{%
\begin{tabular}{llrrrr}
\toprule
\multicolumn{6}{c}{\textbf{\ct}} \\
 &  &  \co &  \fmn &  \mn &  \sv        \\
 \midrule
0 & \textbf{Rob} &     \bff{44.92} &   \bff{76.89} &  \bff{88.11} & \bff{58.34} \\
  & \textbf{St} &     35.76 &   67.15 &  64.17 & 36.67 \\
  \cmidrule{2-6}
1 & \textbf{Rob} &     \bff{61.92} &   \bff{88.78} &  \bff{98.25} & \bff{91.52} \\
  & \textbf{St} &     58.76 &   86.65 &  98.04 & 90.88 \\
  \cmidrule{2-6}
2 & \textbf{Rob} &     \bff{78.85} &   \bff{93.22} &  \bff{99.24} & 96.51 \\
  & \textbf{St} &     78.34 &   92.15 &  99.23 & \bff{96.62} \\
\bottomrule
\end{tabular}
\begin{tabular}{rrrr}
\toprule
\multicolumn{4}{c}{\textbf{\co}} \\
\ct &  \fmn &  \mn &  \sv        \\
\midrule
74.47 &   \tbf{84.85} &  \tbf{94.96} & \tbf{70.61} \\
 \tbf{80.18} &   76.10 &  79.46 & 55.60 \\ \midrule
 85.67 &   \tbf{89.22} &  \tbf{98.33} & \tbf{91.34} \\
  \tbf{87.80} &   88.65 &  97.75 & 91.12 \\ \midrule
  94.82 &   \tbf{92.58} &  \tbf{99.24} & \tbf{96.63} \\
  \tbf{95.20} &   91.78 &  99.22 & 96.60 \\
\bottomrule
\end{tabular}
}
\caption{Transferability of a ResNet50 pretrained on CIFAR-10 (right) and CIFAR-100 (left) in terms of percentage accuracy.}
 \label{tab:transfer10_ext}
\end{table*}

\begin{table*}[!htbp]
\centering
\resizebox{1.0\textwidth}{!}{%
\begin{tabular}{llrrrrrrrrrrr}
\toprule
\textbf{\im} &  &   \ct &  \co &  \fmn &  \mn &  \sv  & \air &  \bir &  \ca &  \dog &  \fl &  \ind \\
\midrule
0 & \rob &  \bff{93.78} &     \bff{77.94} &   \bff{90.09} &  \bff{98.03} & \bff{76.90} & 33.81 &          35.91 & 40.47 & 66.25 &    93.15 &   63.06 \\
  & \sta &  84.72 &     64.48 &   86.38 &  93.91 & 50.46 & \bff{36.72} & \bff{53.58} & \bff{50.12} & \bff{88.39} &    \bff{95.48} &   \bff{72.84} \\
\cmidrule(r){2-13}
1 & \rob &  \bff{94.04} &     \bff{79.01} &   92.53 &  \bff{98.59} & \bff{92.24} & \bff{37.89} &          \bff{28.98} & 57.82 & 65.96 &    94.62 &   63.73 \\
  & \sta &  91.47 &     75.13 &   \bff{92.63} &  98.44 & 91.30 & 37.83 &          24.04 & \bff{64.20} & \bff{78.73} &    \bff{96.21} &   \bff{67.24} \\
\cmidrule(r){2-13}
2 & \rob &  \bff{97.99} &     \bff{87.31} &   \bff{95.51} &  \bff{99.47} & \bff{96.89} & 67.15 &          51.53 & 87.51 & 76.22 &    98.90 &   72.09 \\
  & \sta &  97.10 &     85.54 &   95.21 &  99.35 & 96.47 & \bff{71.32} & \bff{64.33} & \bff{89.38} & \bff{83.34} &    \bff{99.27} &   \bff{76.87} \\
\bottomrule
\end{tabular}
}
\caption{Transferability of a ResNet50 pretrained on ImageNet in terms of percentage  accuracy.}
 \label{tab:transferImage_ext}
\end{table*}

\begin{table*}[htbp]
\centering
\begin{tabular}{lrrrr}
\toprule
\multicolumn{5}{c}{\textbf{\ct}} \\
 &  \co &  \fmn &  \mn &  \sv        \\
\midrule
0    &      9.16 &    9.74 &  23.94 & 21.68 \\
1    &      3.16 &    2.13 &   0.21 &  0.65 \\
2    &      0.51 &    1.07 &   0.01 & -0.10 \\
\bottomrule
\end{tabular}
\begin{tabular}{lrrrr}
\toprule
\multicolumn{4}{c}{\textbf{\co}} \\
\ct &  \fmn &  \mn &  \sv        \\
\midrule
  -5.71 &    8.75 &  15.50 & 15.00 \\
-2.12 &    0.57 &   0.58 &  0.22 \\
-0.11 &    0.80 &   0.02 &  0.03 \\
\bottomrule
\end{tabular}
\caption{Accuracy \% difference between the robust and standard model tranfering from CIFAR-10 (left) and CIFAR-100 (right).}
 \label{tab:diffs}
\end{table*}

\begin{table*}[htbp]
\centering
\begin{tabular}{lrrrrr}
\toprule
\tbf{\im} &  \ct &  \co &  \fmn &  \mn &  \sv  \\
\midrule
0    &   9.06 &     13.46 &    3.71 &   4.12 & 26.44 \\
1    &   2.57 &      3.88 &   -0.10 &   0.15 &  0.94 \\
2    &   0.89 &      1.77 &    0.30 &   0.12 &  0.42 \\
\bottomrule
\end{tabular}

\caption{Accuracy \% difference between the robust and standard model transferring from ImageNet.}
 \label{tab:transferImageSimplediff}
\end{table*}

\begin{table*}[htbp]
\centering
\begin{tabular}{lrrrrrr}
\toprule
\textbf{IMG}  &  \air &  \bir &  \ca &  \dog &  \fl &  \ind \\
\midrule
0    &     -2.91 &         -17.67 & -9.65 & -22.14 &    -2.32 &   -9.78 \\
1    &      0.06 &           4.95 & -6.38 & -12.77 &    -1.59 &   -3.51 \\
2    &     -4.17 &         -12.79 & -1.87 &  -7.12 &    -0.37 &   -4.78 \\
\bottomrule
\end{tabular}

\caption{Accuracy \% difference between the robust and standard model transferring from ImageNet on fine-grained datasets.}
 \label{tab:transferImagediff}
\end{table*}

\subsection{Architecture impact}
We report here a comparison of transfering performance using two different architectures namely ResNet50 and ResNet18, trained on CIFAR-100, to assess the impact of the network capacity. 
It is noticeable that with the more complex network (ResNet50) the gap is reduced in cases where the standard model is better and it is increased in cases where the robust one is better.

\begin{table*}[htbp]
\centering
\begin{tabular}{llrrrr}
\toprule
\multicolumn{6}{c}{\textbf{ResNet50}} \\
 & \tbf{\co} &\ct &  \fmn &  \mn &  \sv        \\
\midrule
0 & \rob &  74.47 &   \tbf{84.85} &  \tbf{94.96} & \tbf{70.61} \\
  & \sta &  \tbf{80.18} &   76.10 &  79.46 & 55.60 \\ \midrule
1 & \rob &    85.67 &   \tbf{89.22} &  \tbf{98.33} & \tbf{91.34} \\
  & \sta &  \tbf{87.80} &   88.65 &  97.75 & 91.12 \\ \midrule
2 & \rob &    94.82 &   \tbf{92.58} &  \tbf{99.24} & \tbf{96.63} \\
  & \sta &  \tbf{95.20} &   91.78 &  99.22 & 96.60 \\
\bottomrule
\end{tabular}
\begin{tabular}{rrrr}
\toprule
\multicolumn{4}{c}{\textbf{ResNet18}} \\
\ct &  \fmn &  \mn &  \sv        \\
\midrule
68.89 &   \tbf{83.40} &  \tbf{94.61} & \tbf{61.08} \\
\tbf{76.50} &   76.30 &  77.98 & 49.32 \\
 \cmidrule{1-4}
82.40 &   \tbf{87.59} &  97.82 & \tbf{89.68} \\
\tbf{85.11} &   86.11 &  \tbf{97.84} & 88.62 \\
 \cmidrule{1-4}
94.59 &   \tbf{92.48} &  \tbf{99.30} & \tbf{96.39} \\
\tbf{95.10} &   92.03 &  99.15 & 96.29 \\
\bottomrule
\end{tabular}
\caption{Performance comparison using different architectures transfering from CIFAR-100.}
\label{tab:arch_diff}
\end{table*}

\begin{table*}[htbp]
\centering
\resizebox{0.95\textwidth}{!}{
\begin{tabular}{lrrrr}
\toprule
\multicolumn{5}{c}{\textbf{ResNet50}} \\
& \ct &  \fmn &  \mn &  \sv        \\
\midrule
 0    & -5.71 &    8.75 &  15.50 & 15.00 \\
 1 & -2.12 &    0.57 &   0.58 &  0.22 \\
 2 & -0.11 &    0.80 &   0.02 &  0.03 \\
\bottomrule
\end{tabular}
\begin{tabular}{lrrrr}
\toprule
\multicolumn{4}{c}{\textbf{ResNet18}} \\
\ct &  \fmn &  \mn &  \sv        \\
\midrule
 -7.61 &    7.10 &  16.63 & 11.76 \\
-2.71 &    1.48 &  -0.02 &  1.06 \\
 -0.51 &    0.45 &   0.15 &  0.10 \\
\bottomrule
\end{tabular}
}
\caption{Performance gap using different pretrained architectures transfering from CIFAR-100.}
\label{tab:arch_diff_gap}
\end{table*}

\end{document}